   \documentclass[12pt]{report}%{book}
   \usepackage{PennDiss}
\usepackage{graphicx}
\usepackage{graphics}
\usepackage{algorithm,amsmath, amsthm, amssymb}
\usepackage{algorithmic}
\usepackage{epstopdf,epsfig}
\usepackage{booktabs}
\usepackage{setspace}
\author{Paramveer S. Dhillon}
\title{Transfer Learning using Feature Selection}
\begin{document}
\pdfpagewidth=8.5in
\pdfpageheight=11.in

%----------------------------------------------------------------------------
%body of document

  \pagenumbering{roman}
  \clearpage

  \title{Transfer Learning using Feature Selection}
  
  \author{Paramveer S. Dhillon}
  \date{April 17, 2009}

  \dept{Computer and Information Science}
  \supervisor{Lyle H. Ungar}
  \groupchair{Rajeev Alur}

  \submitdate{April 2009}
  \copyrightyear{2009}

  \beforepreface
  
   \clearpage
  %\copyrighttrue
  %\copyrightpage %generates copyright page
 
%%Uncomment from here on dh-p 
  \tableofcontents

  \tablespagetrue
  \listoftables
  \figurespagetrue
  \listoffigures

	\prefacesection{Acknowledgements}
  
	I would like to thank Prof. Lyle Ungar for advising work on this thesis, as well as Prof. Ben Taskar (CIS, University of Pennsylvania) and Prof. Dean Foster (Statistics, University of Pennsylvania) for serving on the thesis committee. Besides, this I would also like to thank Prof. Martha Palmer, University of Colorado (Boulder) U.S.A for providing the Word Sense Disambiguation data, and Prof. Dana Pe'er, Columbia University, New York City U.S.A for providing the Yeast dataset. Besides this I would also like to thank Brian Tomasik, Computer Science Department, Swarthmore College, PA, U.S.A for providing help with some of the experiments for MIC.
  \clearpage
  %\abstractp
  \prefacesection{Abstract}
 We present three related ways of using Transfer Learning to improve
 feature selection.  The three methods address different problems, and hence share different kinds of
 information between tasks or feature classes, but all three are based
 on the information theoretic Minimum Description Length (MDL)
 principle and share the same underlying Bayesian interpretation.  The
 first method, MIC,  applies when predictive models are to be built
 simultaneously for multiple tasks (``simultaneous transfer'') that share the same set of
 features.  MIC allows each feature to be added to none, some, or all
 of the task models and is most beneficial for selecting a small set
 of predictive features from a large pool of features, as is common in
 genomic and biological datasets.  Our second method, TPC (Three
 Part Coding), uses a similar methodology for the case when the features
 can be divided into feature classes. Our third method,
 Transfer-TPC, addresses the ``sequential transfer'' problem in which the task to which
 we want to transfer knowledge may not be known in advance and may
 have different amounts of data than the other tasks. Transfer-TPC is
 most beneficial when we want to transfer knowledge between tasks
 which have unequal amounts of labeled data, for example the data for
 disambiguating the senses of different verbs.  We demonstrate the
 effectiveness of these approaches with experimental results on real
 world data pertaining to genomics and to Word Sense Disambiguation
 (WSD).

  \clearpage
  \pagenumbering{arabic}
\chapter{ Introduction}

Classical supervised learning algorithms use a set of feature-label
pairs to learn mappings from the features to the associated
labels. They generally do this by considering each classification task
(each possible label) in isolation and learning a model for that task.
Learning models independently for different tasks often works well,
 but when the labeled data is limited and
expensive to obtain, an attractive alternative is to build shared
models for multiple related tasks. For example,  when one is trying to
predict a set of related responses (``tasks"), be they multiple
clinical outcomes for patients or growth rates for yeast strains under different
conditions, it may be possible to ``borrow
strength'' by sharing information between the models for the different
responses. Inductive transfer can be particularly valuable
 when we have disproportionate amount of labeled data for
``similar tasks''. In such a case, if we build separate models for
each task, then we often get poor predictive accuracies on tasks
which have little data. Transfer learning can potentially be used 
to share information from the tasks with more labeled data to 
``similar'' tasks with less data,  significantly boosting their predictive accuracies.

Transfer learning has been widely used~\cite{Caruana97multitasklearning,ando2005flp,NIPS2008,argyriou,bendavid,koller07,rainangkoller}, 
but generally for determining a shared latent space between
tasks, and not for feature selection.  Our contribution is to present
three models for doing transfer learning that focus on feature
selection. Each of the three models is best suited for a different
problem structure.

The problem of disambiguating word senses based on their context
illustrates the three different types of applications. 

Firstly, each observation of a word (e.g. the sentence containing
the verb ``fire'') is associated with multiple labels corresponding to
each of the different possible meanings (E.g., for firing a person,
firing a gun, firing off a note, etc.)  Rather building separate
models for each each sense (``Is this word sense 1 or not?,'' ``Is
this word sense 2 or not'', etc.), we can note that features that are
useful for predicting one sense are likely to be useful for predicting
the other senses (perhaps with a coefficient of different sign.) 

Secondly, when predicting whether a word has a given sense, we can
group the features derived from its context into different
classes. For example, there are features that characterize the
specific words before and after the target word, features based on the
part of speech labels of those words, and features characterizing the
topic of the document that the ambiguous word is in.  We can
``transfer knowledge'' between the features (not the tasks!), but
noting that when one feature is selected from class, then other
features are more likely to be selected from the same class.

Finally, when predicting whether a word has a given sense, one might
make use of the fact that models for predicting synonyms of that word
are likely to share many of the same features.  I.e., a model for
disambiguating one sense of ``discharge'' is likely to use many of the same
features as one for disambiguating the sense of ``fire'' which is its synonym.

We address all three problems using penalized regression, where linear
or logistic regression models of the form $y = x \beta$ are learned
such that the coefficients (weights) $\beta$ minimize a penalized
likelihood such as

$$ || Y - \hat{Y} ||_2 + \lambda || \beta||_0$$

We use an $\ell_0$ norm on $\beta$ (the number of nonzero
coefficients) to encourage sparse solutions and, critically, we use
information theory to pick the penalty $\lambda$ in a way that
implements the transfer learning.

We can broadly divide the above three problems into two categories. We
address the first two problems using ``simultaneous transfer:''
training data for all the tasks or feature classes are assumed to be
present before learning.  We then select a ``joint'' set of features
shared across the related tasks or feature sets.  We call the
information theoretic penalty used in feature selection MIC (Multiple
Inclusion Criterion) and TPC (Three Part Coding) for the multi-task
and multi-feature class problems, respectively.  We address the third
problem, transfering between different tasks which do not share
observations, as in the case of different words using ``sequential
transfer:'' I.e., we assume that models for some tasks have been
learned and are then used to aid feature selection in building a model
for a new task.  We call the method used for this problem as ``Transfer-TPC.''

We now describe each of these methods (MIC, TPC, and Transfer-TPC'')
in slightly more detail. 

MIC addresses the classic multi-task learning problem \cite{Caruana97multitasklearning}
where each observation is associated with multiple tasks
(a.k.a. multiple labels or multiple responses, {\bf Y}), and allows
each feature to be added to none, some, all of the tasks and is most
beneficial for selecting a small set of predictive features from a
large pool of features.  For example, the tasks can be different
senses of a word, to be predicted from the word context or different
phenotypes (human diseases or yeast growth rates) to be predicted from
a set of gene expression values.

\begin{figure}\label{featClasses}
 
      \center{\epsfxsize=12cm
      \epsffile{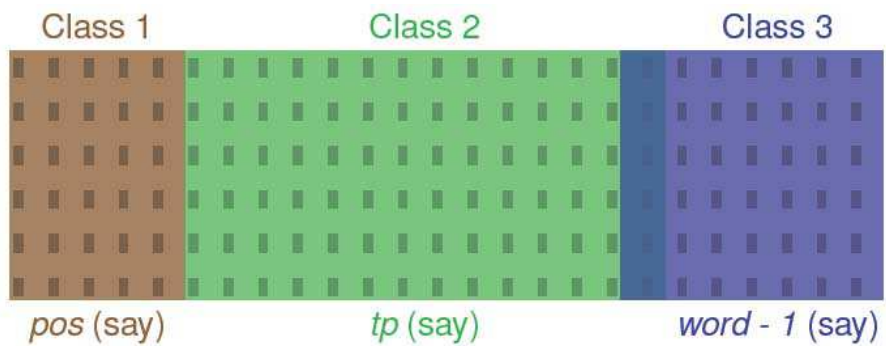}}
\vspace{-2mm}
\caption {$Feature\;Classes\; in\; data$ ($tp~\longmapsto~topic\; of\; the\; document$,  $pos~\longmapsto~part\; of\; speech$, $word-1~\longmapsto~previous\; word$)}
\vspace{-0.4cm} 
\end{figure}

Our second approach, TPC (Three Part Coding), is extremely similar to
MIC, but applies when the features can be divided into feature
classes. Feature classes are pervasive in real data as show in
Fig.~\ref{featClasses}. For example, in gene expression data, the
genes that serve as features may be grouped into classes based on
their membership in gene families or pathways. When doing word sense
disambiguation or named entity extraction, features fall into classes
including adjacent words, their parts of speech, and the topic and
venue of the document the word is in.  When predictive features occur
predominantly in a small number of feature classes, TPC significantly
improves feature selection over naive methods which do not account for
the classes. TPC does not expect the data to have multiple responses,
rather it assume features are shared within classes as opposed to MIC
where they are shared across tasks. The two methods could, of course,
be used together.

MIC tends to include a given feature into more and more tasks as by
doing so the cost of that feature becomes ``cheap'', as explained
below.  TPC tends to include more and more features from a single
feature class as the cost of adding subsequent features from a feature
class is less. They differ slightly in their details due to different
assumptions about the correlation structure of features and responses,
but are otherwise effectively identical.

Transfer-TPC, which uses ``sequential transfer'' from a set of already modeled
``similar'' tasks to guide feature selection on a new task, is somewhat different from
classic multi-task learning methods, in that different feature values and different amounts
of data are available for the different tasks. Transfer-TPC is most beneficial when
we want to transfer knowledge between tasks which have unequal amounts
of labeled data. For example, the VerbNet dataset has roughly six
times more data for one sense of the word ``kill'' than for the
distributionally similar senses of other words like ``arrest'' and
``capture''. In such cases, we can transfer knowledge between these
similar senses of words to facilitate learning predictive models for
the rarer word senses. Transfer-TPC gives significant improvement in performance
in all cases; though the gain in predictive performance is more pronounced when the test task has lesser amount of data than the train tasks, as we demonstrate in Section~\ref{TransTPC}.

Our models use $\ell_0$ penalty instead of the $\ell_1 / \ell_2$
penalty~\cite{argyriou,obozinski} to induce sparsity and select
features. The exact $\ell_0$ penalty requires subset selection, known
to be NP-hard \cite{natarajan1995sas}, but a close solution can be
found by stepwise search. Although approximate, stepwise $\ell_0$
methods generally yield sparser models than exact $\ell_1$ methods
\cite{defensel0}. Moreover, they allow for more flexible choice of
penalties, as we illustrate later in the thesis. All the three models
use information theoretic Minimum Description Length (MDL) principle
\cite{Risannen} to derive an efficient coding scheme for stepwise
regression.

The rest of the thesis is organized as follows. In next chapter we
review relevant previous work. In Chapter 3 we provide background on
basic feature selection methods and the MDL principle. Then in Chapter
4 we provide the general methodology used by all our models. In
Chapters 5, 6 and 7 we describe the MIC, TPC and Transfer-TPC models
in detail, and also show experimental results on real and synthetic
data. In Chapter 8, we give a discussion of all the three models and
show some connections among the three models. We conclude in Chapter 9
by providing a brief summary.

\chapter{ Related Work}

``Multi-Task Learning" or ``Transfer Learning" has been studied
extensively
\cite{Caruana97multitasklearning,ando2005flp,NIPS2008,argyriou,bendavid,koller07,rainangkoller}
in literature. To give a couple examples: \cite{ando2005flp} do joint
empirical risk minimization and treat the multi-response problem by
introducing a low-dimensional subspace which is common to all the
response variables. \cite{rainangkoller} construct a multivariate
Gaussian prior with a full covariance matrix for a set of ``similar"
supervised learning tasks and then use semidefinite programming (SDP)
to combine these estimates and learn a good prior for the current
learning task. \cite{koller07} use the concept of meta-features; they
learn meta-priors and feature weights from a set of similar prediction
tasks using convex optimization.  Some traditional methods such as
neural networks also share parameters between the different
tasks~\cite{Caruana97multitasklearning}.

However, none of the above methods do feature selection. This limits
their applicability in domains such as computational biology (e.g.,
genomics) and language (e.g., Word Sense Disambiguation) \cite{WSD06}
where often only a handful of the thousands of potential features are
predictive and feature selection is very important.  There has been a
small amount of work which does feature selection for multi-task
learning~\cite{argyriou,obozinski}. Both these papers use an $\ell_2$
penalty over coefficients for all tasks associated with a single
feature, combined with an $\ell_1$ penalty over features; this tends
to put each feature into either all or none of the task models.
\cite{argyriou} use this mixed norm ($\ell_1 / \ell_2$) approach for
multi-task feature selection and show that the general subspace
selection problem can be formulated as an optimization problem
involving the trace norm. \cite{obozinski} also use a $\ell_1 /
\ell_2$ block-norm regularization, but they focus on the case where
the trace norm is not required and instead use a homotopy-based
approach to evaluate the entire regularization path efficiently
\cite{lars}.

\chapter{ Background }

Standard feature selection methods for supervised learning assume a
setting consisting of {\it n} observations and a fixed number of {\it
  m} candidate features. The goal of feature selection is to select
the feature subset that will lead to a model with least prediction
error on test set. For many prediction tasks only a small fraction of
the total {\it m} features are beneficial, so good feature selection
methods can give large improvement in predictive accuracy \cite
{nips03}.

The state of the art feature selection methods use either $\ell_0$ or
$\ell_1$ penalty on the coefficients. $\ell_1$ penalty methods such as Lasso
\cite{Lasso} and its variants \cite{fusedlasso,grouplasso}, being
convex, can be solved by optimization and give guaranteed optimal
solutions \cite{lars}. On the other hand, $\ell_0$ penalty methods
require an explicit search through the feature space (as in stepwise,
stagewise and streamwise regression), but have the advantage that they
allow the use of theory to select regularization penalties. As such,
they avoid the usual cross validation used in $\ell_1$ methods, and they
can be easily extended to select penalties in more complex settings as
in this thesis.

The most common of these $\ell_0$ penalty methods is stepwise feature
selection. It is an iterative procedure in which at each step all
features are tested at each iteration, and the best feature is
selected and added to the model. The stepwise search terminates when
either all of the {\it m} candidate features have been added to the
model, or none of the remaining features are beneficial to the model,
according to some measure such as a p-value threshold.

Another, recent method of interest is streamwise feature selection
\cite{zhou:06:jmlr} (SFS), which is a greedy online method. In this
method each feature is evaluated for addition to the model only once
and if the reduction in prediction error resulting from adding the
feature to the model is more than an ``adaptively adjusted" threshold
then that feature is added to the model. It contrasts with the
``batch" methods as Support Vector Machines (SVMs), neural nets
etc. which require having all features in advance. SFS is somewhat
similar to an alternate class of feature selection methods that
control the False Discovery Rate (FDR)\cite{Benjamini}, and scales
well to very large feature sets.

\chapter{General Methodology}
In this chapter we describe the basic methodology that all our three models share, i.e. use  MDL (Minimum Description Length) based coding schemes.

All our three models use a Minimum Description Length
(MDL) \cite{Risannen} based coding scheme, which we explain in the, to specify another 
penalized likelihood method.

In general, penalized likelihood methods aim to minimize an objective function of the form
\begin{equation}
\label{penlikelihood}
\text{score} = -2\log(\text{likelihood of {\bf Y} given {\bf X}}) + F \times q,
\end{equation}
where $q$ is the current number of features in the model. Various penalties $F$ have been proposed, including $F=2$, corresponding to AIC (Akaike Information Criterion), $F=\log n$, corresponding to BIC (Bayesian
Information Criterion), and $F = 2 \log m$, corresponding to RIC (Risk Inflation Criterion---similar to
a ``Bonferroni correction")~\cite{FosterGeorge}.

The penalties for these methods are summarized in the Table~\ref{tab:PenIC}.

\begin{table} [htbp]
\setlength{\abovecaptionskip}{0pt} 
\setlength{\belowcaptionskip}{0pt}
\caption{Penalties for different Information Criterion methods.}
\begin{small}
\label{tab:PenIC}
\vskip 0.02in
\begin{center}
\begin{tabular}{lccc}
\hline
Name & Penalty & Assumption\\
\hline
\small
Akaike Information Criterion(AIC) & 2 & - \\
Bayesian Information Criterion (BIC) & log(n) & n$\gg$p\\
Risk Inflation Criterion (RIC) & 2log(p) & p$\gg$n\\
\hline
\end{tabular}
\end{center}
\end{small}
\vskip -0.02in
\end{table}

Each of these penalties can be interpreted within the framework of the Minimum Description Length (MDL) principle \cite{Risannen}. MDL envisions a ``sender," who knows {\bf X} and {\bf Y}, and a ``receiver," who knows only {\bf X}. In order to transmit {\bf Y} using as few bits as possible, the sender encodes not the raw {\bf Y} matrix but instead a model for {\bf Y} given {\bf X}, followed  by the residuals of {\bf Y} about that model. The length $S$ of this message, in bits, is called the {\it description length} and is the sum of two components. The first is $S_E$, the number of bits for encoding the residual errors, which according to standard MDL is given by the negative log-likelihood of the data given the model; this can be identified with the first term of Equation~\ref{penlikelihood}. The second component, $S_M$, is the number of bits used to describe the model itself and can be seen as corresponding to the second term of Equation~\ref{penlikelihood}.

For MIC, we use the term {\it total description length} (TDL) to denote the combined length of the message for all {\it h}  tasks and hence we select features for the $h$
responses (tasks)\footnote{The notion of ``task'' in this section is a separate response vector; and it is different than the general notion of ``task'' in transfer learning (For e.g. in Transfer-TPC), where it may not necessarily mean a separate responce vector} simultaneously to minimize $S$. Thus, when we evaluate a feature for addition into the
model, we want to maximize the reduction of TDL $\Delta S^k$ incurred by adding
that feature to a subset $k$ of the $h$ tasks $(1 \leq k \leq h)$:
\begin{equation}
\nonumber
\Delta S^k = \Delta S_E^k - \Delta S_M^k
\end{equation}
where $\Delta S_E^k > 0$ is the reduction in residual-error coding cost due to the data likelihood increase given the new feature, and $\Delta S_M^k > 0$ is the increase in model cost to encode the new feature.\footnote{ $\Delta S_E^k$ is always greater than zero, because even a spurious feature will slightly increase the data likelihood.}

As will be seen in Section \ref{coding-schemes}, MIC's model cost i.e. ($\Delta S_M$) includes a component for coding feature coefficients that resembles the AIC or BIC penalty, plus a component for specifying which features are present in the model that resembles the RIC penalty.

In case of TPC and Transfer-TPC the definition of the term {\it total description length} (TDL) is a bit different and over there it is just the length of the message for the single response (task) and $\Delta S_M$ consists of; the class of the feature being added, which feature in the class, and what is its coefficient.

\chapter{ Model 1: MIC (Multiple Inclusion Criterion) }
In this chapter we explain MIC, which is a model for transfer/ multi -task learning, and does ``simultaneous transfer'' (joint feature selection) for multiple related tasks which share the same set of features. It uses MDL (Minimum Description Length) principle to derive an efficient coding scheme for multi -task stepwise regression.
Firstly, we describe the notation used and provide a basic overviwe of MIC. Then we describe the coding schemes used in MIC and provide a comparison of various MIC coding schemes.

\section{Notation Used}

The symbols used throughout this section are defined in the
Table~\ref{tab:Symbols}. All the values in the table are given by data
except $m^*$, which is unknown.

\begin{table} [thbp]
\setlength{\abovecaptionskip}{1pt} 
\setlength{\belowcaptionskip}{1pt}
\caption{Symbols used and their definitions.}
\centering
\label{tab:Symbols}
\begin{center}
\begin{tabular}{c|l}
\toprule
Symbol & Meaning\\
\midrule
\small
$n$ & Total number of observations\\
$m$ &  Number of candidate features\\
$m^*$ & Number of beneficial features\\
%$q$ &  Number of features currently included \\
%& in the model\\
$h$ & Total number of tasks\\
$k$ & Number of tasks into which a feature \\
& \ \ \  has been added\\
%$i$ & Index of task\\ <-- not used
$j$ & Index of feature\\
$i$ & Index of observation\\
\bottomrule
\end{tabular}
\end{center}
\end{table}

Thus, we have an $n \times h$ response matrix {\bf Y}, with a shared $n \times m$ a feature matrix~{\bf X}.

\section{Coding Schemes used in MIC}\label{coding-schemes}
In this section we describe the coding scheme used by MIC for the general case in which features can be added to a subset of tasks but the tasks share strength. In next section we explore the special cases in which a feature is added to all tasks or none and features are added independently to each task (i.e., no transfer). 

\subsection{Code $\Delta S_{jE}^k$}\label{regression-code}

Let {\bf E} be the residual error matrix:
\begin{equation}
\nonumber
{\bf E = Y - \hat{Y}},
\end{equation}
where {\bf Y} and ${\bf \hat{Y}}$ are the $n \times h$ response and prediction
matrices, respectively.

$\Delta S_{jE}^k$ is the decrease in negative log-likelihood that results from adding feature $j$ to some subset $k$ of the $h$ tasks.
If all the tasks were independent, then $\Delta S_{jE}^k$ would simply be the sum of
the changes in negative log-likelihood for each of the $h$ models separately. However, we may want our model to allow for nonzero covariance among the tasks. This is particularly true for stepwise regression, because in the first iterations of a stepwise algorithm, the effects of features not present in the model show up as part of the ``noise" error term, and if two tasks share a feature not yet in the model, the portion of the error term due to that feature will be the same.

Thus, letting $e_i$, $i = 1, 2, \ldots, n$, denote the error for the $i^\text{th}$ row of {\bf E}, we assume $e_i \stackrel{\text{\tiny i.i.d.}}{\sim} \mathcal{N}(0,\Sigma)$, with $\Sigma$ an $h \times h$ covariance matrix. In other words,
\begin{equation}
\nonumber
P(e_i)= \frac{1}{\sqrt{ (2\pi)^h |\Sigma|}} \exp \left( -\frac{1}{2} e_i^T \Sigma^{-1}e_i \right)
\end{equation}
in which $(\cdot)^T$, $(\cdot)^{-1}$, and $|\cdot|$ are the matrix transpose, inverse, and determinant, respectively. Therefore,
\begin{equation}\label{negloglike}
\begin{split}
S_E&=-\log \prod_{i=1}^n P(e_i)\\
& = \frac{n}{2} \log{ \left( (2\pi)^h |\Sigma| \right) } + \frac{1}{2 \ln 2}\sum_{\nu=1}^n e_i^T \Sigma^{-1}e_i,\\
\end{split}
\end{equation}
where the $\frac{1}{\ln 2}$ factor appears because we use logarithm base 2 (here and throughout the remainder of the paper). Note that the superscript $k$ in $\Delta S_{jE}^k$ indicates that the reduction is incurred by adding a new feature to $k$ tasks, but the calculation $\Delta S_{jE}^k$ is over all {\it h} tasks; i.e., the whole residual error {\bf E} is taken into account.

\subsection{Code $\Delta S_{jM}^k$}

To describe $\Delta S_{jM}^k$ when a feature is added, MIC uses a
three part coding scheme: 
\begin{equation}
\nonumber
\Delta S_{jM}^k = \ell_I + \ell_H + \ell_\theta,
\end{equation}
where $\ell_I$ is the number of bits needed to describe which feature is being added, $\ell_H$ is the cost of specifying the subset of $k$ of the $h$ task models in which to include the feature, and $\ell_\theta$ is the description length of the $k$ nonzero feature coefficients. We now consider different coding schemes for $\ell_I$, $\ell_H$, and $\ell_\theta$.

\paragraph{Code $\ell_I$}

For most data and feature sets, little is known \textit{a priori} about which features will be beneficial.\footnote{Following \cite{zhou2006sfs}, we define a ``beneficial" feature as one which, if added to the model, would reduce error on a hypothetical infinite test set.} We therefore assume that if a feature $x_j$ is beneficial, its
index $j$ is uniformly distributed over $\{1, 2, \ldots, m\}$. This implies $\ell_I = \log m$ bits to encode the index, reminiscent of the RIC penalty for equation \eqref{penlikelihood}.

RIC often uses no bits to code the coefficients of the features that are added, based on the assumption that $m$ is so large that the $\log m$ term dominates. This assumption is not valid in the multiple response setting, where the number of models $h$ could be large. If a feature is added to $k$ of the $h$ tasks, the cost of encoding the $k$ coefficients may be a major part of the cost.  We describe the cost to code a coefficient below.

\paragraph{Code $\ell_\theta$}~\label{AICcode}
This term corresponds to the number of bits required to code the value
of the coefficient of each feature. We could use either AIC or the
more conservative BIC to code the coefficients. As explained below, we use $2$
bits for each coefficient, similar to AIC.

Given a model, MDL chooses the values of the coefficients that maximize the likelihood of the data. \cite{Risannen2} proposes approximating $\theta$, the Maximum Likelihood Estimate (MLE), using a grid resolved to the nearest standard
error. That is, instead of specifying $\theta$, we encode a rounded-integer value of $\theta$'s z-score $\hat{z}$, where $\theta = \theta_0 + \hat{z} \ \text{SE}(\theta)$, with $\theta_0$ being the default, null-hypothesis value (here, 0) and SE($\theta$) being the standard error of $\theta$.

We assume a
``universal prior" distribution for $\hat{z}$, in which half of the probability is
devoted to the null value $\theta_0$ and the other half is concentrated near $\theta_0$ and decays
slowly. In particular, for $\theta \neq \theta_0$, the coding
cost is 2 + $\log^+\left|\hat{z}\right|$
+ $2 \log^+\log^+\left|\hat{z}\right|$ bits. This prior
distribution makes sense in hard problems of
feature selection where beneficial features are just marginally
significant. Since $\hat{z}$ is quite small in such hard problems, the
2 bits will dominate the other two terms. In fact, we simply assume $l_\theta = 2.$

\paragraph{Code $\ell_H$}

In order to specify the subset of task models that include a given feature, we encode two pieces of information: First, how many $k$ of the tasks have the feature? Second, which subset of $k$ tasks are those?

One way to encode $k$ is to use $\log h$ bits to specify an integer in $\{1, 2, \ldots, h\}$; this implicitly corresponds to a uniform prior distribution on $k$. However, since we generally expect that smaller values of $k$ are more likely, we instead use coding lengths inspired by the ``idealized universal code for the integers" of \cite{elias1975} and \cite{rissanen1983}: The cost to code $k$ is $\log^* k + c_h$, where $\log^* k = \log k + \log \log k + \log \log \log k + \ldots$ so long as the terms remain positive, and $c_h$ is the constant required to normalize the implied probability distribution over $\{1, 2, \ldots, h\}$. $c_\infty \approx 2.865$ \cite{rissanen1983}, but for $h \in \{5, \ldots, 1000\}$, $c_h \approx 1$.

Given $k$, there are $h \choose k$ subsets of tasks to which we can refer, which we can do by coding the index with $\log {h \choose k}$ bits.

Thus, in total, we have
\begin{equation}
\label{lh_code}
\nonumber
\ell_H = \log^* k + c_h + \log {h \choose k}.
\end{equation} 

\section{Comparison of the Coding Schemes}\label{comparison}

The preceding discussion outlined a coding scheme for what we might call ``Partially Dependent MIC," or ``Partial MIC," in which models for different tasks can share some or all features.

As suggested earlier, we can also consider a ``Fully Dependent MIC," or ``Full MIC,"  scheme in which each feature is shared across all or none of the task models. This amounts to a restricted Partial MIC in which $k=0$ or $k=h$ for each feature. The advantage comes in not needing to specify the subset of tasks used, saving $\ell_H$ bits for each feature in the model; however, Full MIC may need to code more coefficient values than Partial MIC.

A third coding scheme is simply to specify each task model in isolation from the others. We call this the ``RIC" approach, because each model pays $\log m$ bits for each feature to code its
index; this is equivalent, up to the base of the logarithm, to the $F = \ln m$ penalty in equation \ref{penlikelihood}. (However, we include an additional cost of $\ell_\theta$ bits to code a coefficient.) If the sum of
the two costs is sufficiently less than the bits saved by the increase of
the data likelihood from adding the feature to the model, 
the feature will be added to the model. RIC assumes that the beneficial features are not significantly shared across tasks.

We compare the relative coding costs under these three coding schemes
for the case where we evaluate a hypothetical feature, $x_j$, that is beneficial
for $k$ tasks and spurious for the remaining  $h - k$ tasks. Suppose that Partial MIC and RIC both add the feature to only the $k$ beneficial tasks, while Full MIC adds it to all $h$ tasks. We assume that if the
feature is added, the three methods save approximately the same number of bits in
encoding residual errors, $\Delta S_E^k$. This would happen if, say, the additional $h-k$ coefficients that Full MIC adds to its models save a negligible number of residual-coding bits (because those features are spurious) and if the estimate for $\Sigma$ is sufficiently diagonal that the negative log-likelihood calculated using \eqref{negloglike} for Partial MIC approximately equals the sum of the negative log-likelihoods that RIC calculates for each response separately.

Table~\ref{tab:compare} shows that RIC and Partial MIC are the best and the second best coding schemes when
$k=1$, and that their difference is on the order of $\log h$. Full MIC and Partial MIC are the best and the second
best coding schemes when $k = h$, and their difference is on the
order of $\log^*h$. Partial MIC is best for $k=\frac{h}{4}$.

\begin{table*} [htbp]
\setlength{\abovecaptionskip}{1pt} 
\setlength{\belowcaptionskip}{1pt}
\caption{Costs in bits for each of the three schemes to code a model with $k=1$, $k=\frac{h}{4}$, and $k=h$ nonzero coefficients. $m \gg h \gg 1$, $\ell_I = \log m$, $\ell_\theta=2$, and for $h \in \{5, \ldots, 1000\}$, $c_h \approx 1$. Examples of these values for $m=2{,}000$ and $h=20$ appear in brackets; the smallest of the costs appears in bold.}
\centering
\label{tab:compare}
\begin{center}
\begin{tabular}{c|cc|cc|cc}
\toprule
\small
$k$ & \multicolumn{2}{c|}{Partial MIC} & \multicolumn{2}{c|}{Full MIC} & \multicolumn{2}{c}{RIC}\\
\midrule
1 & $\log m + c_h + \log h + 2$ & [18.4]  & $\log m + 2 h$ & [51.0] & $\log m+ 2$ & {\bf [13.0]}\\
$\frac{h}{4}$ & $\log m + \log^*\left (\frac{h}{4} \right) + c_h + \log {h \choose h/4} + \frac{h}{2}$ & {\bf [39.8]} & $\log m + 2 h$ & [51.0] & $\frac{h}{4} \log m + \frac{h}{2}$ & [64.8]\\
$h$ & $\log m + \log^* h + c_h + 2 h$ & [59.7] & $\log m + 2 h$ & {\bf [51.0]}  & $h \log m + 2 h$ & [259.3]\\
\bottomrule
\end{tabular}
\end{center}
\end{table*}

\section{Stepwise Search Method}

To search for a model that approximately minimizes TDL, we use a modified greedy stepwise-search algorithm. For each feature, we evaluate the change in TDL that would result from adding that feature to the model with the optimal number of associated tasks. We add the best feature and then recompute changes in TDL for the remaining features. This continues until there are no more features that would reduce TDL. The number of evaluations of features is thus $\mathcal{O}(m m^*)$, where $m^*$ is the number of features eventually added.

To select the optimal number $k$ of task models in which to include a
given feature, we again use a stepwise-style search. In this case, we
evaluate the reduction in TDL that would result from adding the
feature to each task, add the feature to the best task, recompute
the reduction in TDL for the remaining tasks, and continue.\footnote{A
  stepwise search that re-evaluates the quality of each task at each
  iteration is necessary because, if we take the covariance matrix
  $\Sigma$ to be nondiagonal, the values of the residuals for one task
  may affect the likelihood of residuals for other tasks. If we take
  $\Sigma$ to be diagonal, as we do in Section
  \ref{Experimental Results}, then an $\mathcal{O}(h)$ search through
  the tasks without re-evaluation suffices. } However, unlike a
normal stepwise search, we continue this process until we have added the
feature to all $h$ task models. The reason for this is two-fold.
First, because we want to borrow strength across tasks, we
need to avoid overlooking cases where the correlation of a feature
with any single task is insufficiently strong to warrant addition, yet
the correlations with all of the tasks are. Second, the $\log
\binom{h}{k}$ term in Partial MIC's coding cost does not increase
monotonically with $k$, so even if adding the feature to an
intermediate number of tasks does not look promising, adding it to all
of them might still be worthwhile.

Thus, for a given feature, we evaluate the description length of the
model $\mathcal{O}(h^2)$ times. Since we need to identify the
optimal $k$ for each feature evaluation, the entire algorithm requires
$\mathcal{O}(h^2 m m^*)$ evaluations of TDL. However, with a few
optimizations, this cost can be reduced with no practical
impact on performance:
\begin{itemize}
\item We can quickly filter out most of the irrelevant features at
  each iteration by evaluating, for each feature, the decrease in
  negative log-likelihood that would result from simply adding it with
  all of its tasks, without doing any subset search. Then we keep only
  the top $t$ features according to this criterion, on which we
  proceed to do the full $\mathcal{O}(h^2)$ search over subsets. We
  use $t = 75$, but we find that as long as $t$ is bigger than, say,
  10 or 20, it makes essentially no impact to the quality of
  results. This reduces the number of model evaluations to
  $\mathcal{O}(m m^* + m^* t h^2)$.
\item We can often short-circuit the $\mathcal{O}(h^2)$ search over
  task subsets by noting that a model with more nonzero coefficients
  always has lower negative log-likelihood than one with fewer nonzero
  coefficients. This allows us to get a lower bound on the description
  length for the current feature for each number $k \in \{1, \ldots,
  h\}$ of nonzero tasks that we might choose as
\begin{equation}\label{boundEqn}
\nonumber
\begin{split}
&(\text{Model cost for other features already in model}) \\
 &+ (\text{negative log-likelihood of $Y$ if we included all $h$ tasks for this feature})\\
& +  (\text{the increase in model cost if we included just $k$ of the tasks}).
\end{split}
\end{equation}
We then need only check those values of $k$ for which \eqref{boundEqn}
is smaller than the best description length for any candidate
feature's best task subset seen so far. In practice, with $h=20$, we
find that evaluating $k$ up to, say, 3 or 6 is usually enough; i.e.,
we typically only need to add $3$ to $6$ tasks in a stepwise manner
before stopping, with a cost of only $3h$ to $6h$.\footnote{If
  $\Sigma$ is diagonal and we do not need to re-evaluate residual
  likelihoods at each iteration, the cost is only $3$ to $6$
  evaluations of description length.}
\end{itemize}

Although we did not attempt to do so, it may be possible to formulate
MIC using a \emph{regularization path}, or \emph{homotopy}, algorithm
of the sort that have become popular for performing $\ell_1$
regularization without the need for cross-validation (e.g.,
\cite{friedman2008rpg}). If possible, this would be significantly
faster than stepwise search.

\section{Experimental Results}\label{Experimental Results}

This section evaluates the MIC approach on three synthetic
datasets, each of which is designed to match the assumptions of, respectively, the Partial MIC, Full MIC, and RIC coding schemes described in Section \ref{comparison}. We also test on two
biological data sets, a Yeast Growth dataset \cite{litvin2009mig},
which consists of real-valued growth measurements of multiple strains of yeast
under different drug conditions, and a Breast Cancer dataset
\cite{vantveer}, which involves predicting prognosis, ER status, and three other 
descriptive variables from gene-expression values for different cell lines.

We compare the three coding schemes of Section \ref{comparison} against two other multitask algorithms: ``AndoZhang" \cite{ando2005flp} and
``BBLasso" \cite{obozinski}, as implemented in the Berkeley Transfer Learning Toolkit
\cite{TLToolkit}. We did not compare MIC with other methods from the
toolkit as they all require the data to have additional
structure, such as {\it meta-features} \cite{koller07,rainangkoller}, or
expect the features to be frequency counts, such as for the Hierarchical
Dirichlet Processes algorithm. Also, none of the neglected methods does feature selection.
 
For AndoZhang we use 5-fold CV to find the best $h$ parameter (the
dimension of the subspace ($\Theta$), not to be confused with $h$ as
we use it in this paper). We tried values in the range $[1, 100]$ as is
done in \cite{ando2005flp}. For MIC, one can use 
either a full or a diagonal covariance matrix estimate. We found that
substantial overfitting can occur when using a full covariance matrix, and
therefore used a diagonal covariance matrix in all experiments presented below.

MIC as presented in Section \ref{regression-code} is a regression algorithm, but AndoZhang and BBLasso are both designed for classification. Therefore, we made each of our responses binary 0/1 values before applying MIC with a regular regression likelihood term. Once the features were selected, however, we used logistic regression applied to just those features to obtain MIC's actual model coefficients.

As noted in Section \ref{regression-code}, MIC's negative log-likelihood term can be computed with an arbitrary $h \times h$ covariance matrix $\Sigma$ among the $h$ tasks. On the data sets in this paper, we found that estimating all $h^2$ entries of $\Sigma$ could lead to overfitting, so we instead took $\Sigma$ to be diagonal. Informal experiments showed that estimating $\Sigma$ as a convex combination of the full and diagonal estimates could also work well.

\subsection{Evaluation on Synthetic Datasets}\label{syn-data-settings}
We created synthetic data according to three separate scenarios---called Partial, Full, and Independent. For each scenario, we generated a matrix of continuous responses as
\begin{equation}
\nonumber
{\bf Y}_{n \times h} = {\bf X}_{n \times m} {\bf \beta}_{m \times h} + \epsilon_{n \times h}
\end{equation}

where $m=2{,}000$ features, $h=20$ responses, and $n=100$ observations.
 Then, to produce binary responses, we set to 1 those response values that were greater than or equal to the average value for their column and set to 0 the rest; this produced a roughly 50-50 split between 1's and 0's. 
Each nonzero entry of $\beta$ was i.i.d. $\mathcal{N}(0,1)$, and entry of $\epsilon$ was i.i.d. $\mathcal{N}(0, 0.1)$, with no covariance among the $\epsilon$ entries for different tasks. Each task had $m^*=4$ beneficial features, i.e., each column of $\beta$ had 4 nonzero entries.

The scenarios differed according to the distribution of the beneficial features in $\beta$.
\begin{itemize}
\item In the Partial scenario, the first feature was shared across all 20 responses, the second was shared across the first 15 responses, the third across the first 10 responses, and the fourth across the first 5 responses. Because each response had four features, those responses ($6-20$) that did not have all of the first four features had other features randomly distributed among the remaining features (5, 6, \ldots, 2000).
\item In the Full scenario, each response shared exactly features $1-4$, with none of features $5-2000$ being part of the model.
\item In the Independent scenario, each response had four random features among ${1, \ldots, 2000}$.
\end{itemize}

For the synthetic data, we report precision and recall to measure the quality of feature selection. This can be done both at a coefficient level (Was each nonzero coefficient in $\beta$ correctly identified as nonzero, and vice versa?) and at an overall feature level (For features with \textit{any} nonzero coefficients, did we correctly identify them as having nonzero coefficients for any of the tasks, and vice versa?). Note that Full MIC and BBLasso always make entire rows of their estimated $\beta$ matrices nonzero and so tend to have larger numbers of nonzero coefficients. Table
\ref{partialTablehIs20} shows the performance of each of the methods
on five instances of the Partial, Full, and Independent synthetic data sets. On the Partial data set, {\it
  Partial MIC} performed the best, closely followed by {\it
  RIC}; on the Full synthetic data, {\it Full MIC} and
{\it Partial MIC} performed equally well; and on the Independent
synthetic data, the {\it RIC} algorithm performed the best closely followed
by {\it Partial MIC}.  It is also worth noting that the best performing
methods tended to have the best precision and recall on coefficient
selection. The performance trends of the three methods are in
consonance with the theory of Section \ref{comparison}.

The table shows that only in one of the three cases does one of these
methods compete with MIC methods. BBLasso on the Full synthetic data
shows comparable performance to the MIC methods, but even in that case
it has a very low feature precision, since it added many more spurious features than the
MIC methods.

% Below from chervil[results_23-Jan-2009_nDat=1]$

\begin{table*}[t]
\caption{Test-set accuracy, precision, and recall of MIC and other methods on 5 instances of various synthetic data sets generated as described in Section \ref{syn-data-settings}. Standard errors are reported over each task; that is, with 5 data sets and 20 tasks per data set, the standard errors represent the sample standard deviation of 100 values divided by $\sqrt{100}$.  {\it Note:} AndoZhang's NA values are due to the fact that it does not explicitly select features.}
%\centering
\begin{tabular}{cccc}
\hline
Method & Test Error & Coefficient & Feature\\ 
 & $\mu\pm\sigma$ & Precision/Recall& Precision/Recall\\
\hline
\multicolumn{4}{c}{Partial Synthetic Dataset}\\
\hline
True Model & $0.07 \pm 0.00$ & $1.00 \pm 0.00 $/$1.00 \pm 0.00 $& $1.00 \pm 0.00 $/$1.00 \pm 0.00 $\\

Partial MIC  &{\bf  0.10 $\pm$ 0.00}& $0.84 \pm 0.02 $/$0.77 \pm 0.02 $& $0.99 \pm 0.01 $/$0.54 \pm 0.05 $\\

Full MIC  & $0.17 \pm 0.01$ & $0.26 \pm 0.01 $/$0.71 \pm 0.03 $& $0.97 \pm 0.02 $/$0.32 \pm 0.03 $\\

Indep. & $0.12 \pm 0.01$ & $0.84 \pm 0.02 $/$0.56 \pm 0.02 $& $0.72 \pm 0.05 $/$0.62 \pm 0.04 $\\

AndoZhang & $0.50 \pm 0.02$ & NA & NA\\

BBLasso & $0.19 \pm 0.01$ & $0.04 \pm 0.00 $/$0.81 \pm 0.02 $& $0.20 \pm 0.03 $/$0.54 \pm 0.01 $\\
\hline
\multicolumn{4}{c}{Full Synthetic Dataset}\\
\hline
True Model  & $0.07 \pm 0.00$& $1.00 \pm 0.00 $/$1.00 \pm 0.00 $& $1.00 \pm 0.00 $/$1.00 \pm 0.00 $\\

Partial MIC  &{\bf  0.08 $\pm$ 0.00}& $0.98 \pm 0.01 $/$1.00 \pm 0.00 $& $0.80 \pm 0.00 $/$1.00 \pm 0.00 $\\

Full MIC  &{\bf 0.08 $\pm$ 0.00}& $0.80 \pm 0.00 $/$1.00 \pm 0.00 $& $0.80 \pm 0.00 $/$1.00 \pm 0.00 $\\

Indep.   & $0.11 \pm 0.01$& $0.86 \pm 0.02 $/$0.63 \pm 0.02 $& $0.36 \pm 0.06 $/$1.00 \pm 0.00 $\\

AndoZhang & $0.45 \pm 0.02$ & NA & NA\\

BBLasso & $0.09 \pm 0.00$& $0.33 \pm 0.03 $/$1.00 \pm 0.00 $& $0.33 \pm 0.17 $/$1.00 \pm 0.00 $\\
\hline
\multicolumn{4}{c}{Independent Synthetic Dataset}\\
\hline
True Model & $0.07 \pm 0.00$& $1.00 \pm 0.00 $/$1.00 \pm 0.00 $& $1.00 \pm 0.00 $/$1.00 \pm 0.00 $\\

Partial MIC  & $0.17 \pm 0.01$& $0.95 \pm 0.01 $/$0.44 \pm 0.02 $& $1.00 \pm 0.00 $/$0.44 \pm 0.02 $\\

Full MIC & $0.36 \pm 0.01$& $0.06 \pm 0.01 $/$0.15 \pm 0.02 $& $1.00 \pm 0.00 $/$0.14 \pm 0.02 $\\

Indep. & {\bf 0.13 $\pm$ 0.01}& $0.84 \pm 0.02 $/$0.58 \pm 0.02 $& $0.83 \pm 0.02 $/$0.58 \pm 0.03 $\\

AndoZhang & $0.49 \pm 0.00$ & NA & NA\\

BBLasso & $0.35 \pm 0.01$& $0.02 \pm 0.00 $/$0.43 \pm 0.02 $& $0.30 \pm 0.05 $/$0.42 \pm 0.06 $\\
\hline
\end{tabular}
\label{partialTablehIs20}
\end{table*}

\subsection{Evaluation on Real Datasets}
This section compares the performance of MIC methods with AndoZhang and BBLasso on a Yeast dataset and Breast Cancer dataset. These are typical of biological datasets in that only a handful of features are predictive from thousands of potential features. This is precisely the case in which MIC outperforms other methods. MIC not only gives better accuracy but does so by choosing fewer features than BBLasso's $\ell_1 / \ell_2$-based approach.     

\subsubsection{Yeast Dataset}
Our Yeast dataset comes from \cite{litvin2009mig}. It consists of real-valued growth measurements of 104 strains of yeast ($n=104$ observations) under 313 drug conditions. In order to make computations faster, we hierarchically clustered these 313 conditions into 20 groups using correlation as the similarity measure. Taking the average of the values in each cluster produced $h=20$ real-valued responses (tasks), which we then binarized into two categories: values at least as big as the average for that response (set to 1) and values below the average (set to 0). The features consisted of 526 markers (binary values indicating major or minor allele) and 6,189 transcript levels in rich media for a total of $m = 6{,}715$ features. 

Table \ref{growthTable20} shows 
test errors from 5-fold CV on this data set. As can be seen from the table,
Partial MIC performs better than BBLasso. BBLasso
overfits substantially, as is shown by 
its large number of nonzero coefficients. We also note that RIC and Full MIC perform slightly worse than Partial MIC, underscoring the point that it is preferable to use a more general MIC coding scheme
compared to Full MIC or RIC. The latter methods have 
strong underlying assumptions, which cannot always
correctly capture sharing across tasks. Like Partial MIC, AndoZhang did well on this data set; however, because the algorithm scales poorly with large numbers of tasks, the computation took 39 days.

\begin{table*}[t]
\caption{Accuracy and number of coefficient and features selected on five folds of CV for the Yeast and Breast Cancer data sets. For the Yeast data, $h=20$, $m=6{,}715$, $n=104$.  For the Breast Cancer data, $h=5$, $m=5{,}000$, $n=100$. Standard errors are over the five CV folds; i.e., they represent (sample standard deviation) / $\sqrt{5}$.  {\it Note:} {\bf These are true cross validation accuracies and no parameters have been tuned on them.} AndoZhang's NA values are due to the fact that it does not explicitly select features. }
\centering
\begin{tabular}{cccccc} 
\hline
Method & Partial MIC  & Full MIC & RIC & AndoZhang & BBLasso\\ 
\midrule
\multicolumn{6}{c}{Yeast Dataset}\\
\midrule
Test error 
 &{\bf  0.38 $\pm$ 0.04}
 & $0.39 \pm 0.04$
  & $0.41 \pm 0.05$
 & $0.39 \pm 0.03$
 & $0.43 \pm 0.03$
\\
Num. coeff. sel. 
  & $22 \pm 4$
 & $64 \pm 4$
 & $9 \pm 1$
 & NA
 & $1268 \pm 279$
 \\
 Num. feat. sel. 
  & $4 \pm 0$
 & $3 \pm 0$
 & $9 \pm 1$
 & NA
 & $63 \pm 14$
 \\
\midrule
\multicolumn{6}{c}{Breast Cancer Dataset}\\
\midrule
Test error 
& {\bf  0.33 $\pm$ 0.08}
 & $0.37 \pm 0.08$
  & $0.36 \pm 0.08$
 & $0.44 \pm 0.03$
 &{\bf 0.33 $\pm$ 0.08}
\\
Num. coeff. sel. 
  & $3 \pm 0$
 & $11 \pm 1$
  & $2 \pm 0$
 & NA
 & $61 \pm 19$
 \\
 Num. feat. sel. 
  & $2 \pm 0$
 & $2 \pm 0$
  & $2 \pm 0$
 & NA
 & $12 \pm 4$
 \\
\bottomrule
\end{tabular}
\label{growthTable20}
\end{table*}

\subsubsection{Breast Cancer Dataset}

Our second data set pertains to Breast Cancer, containing data from five of the seven data sets used in
\cite{vantveer}. It contains $1{,}171$ observations for $22{,}268$
RMA-normalized gene-expression values. We considered five associated
responses (tasks); two were binary---prognosis (``good" or ``poor") and ER
status (``positive" or ``negative")---and three were not---age (in
years), tumor size (in mm), and grade (1, 2, or 3). We binarized the
three non-binary responses into two categories: Response values at
least as high as the average, and values below the average. Finally we
scaled the dataset down to $n=100$ and $m=5{,}000$ (the 5{,}000 features
with the highest variance), to save computational resources. Table
\ref{growthTable20} shows %training and
test errors from 5-fold CV on
this data set. As is clear from the table, Partial MIC and BBLasso are
the best methods here. But as was the case with other datasets,
BBLasso puts in more features, which is undesirable in domains
(like biology and medicine) where simpler and hence more interpretable model
are sought.

\chapter{Model 2: TPC (Three Part Coding)}\label{TPCChapter}
In this chapter we describe our second model, TPC. As mentioned earlier TPC is quite similar to MIC, and extends the concept of ``joint'' feature selection to the case when the feature matrix has structure i.e. the features are compartmentalized into feature classes. 

The concept of feature classes is very similar to the concept of {\it
  meta - features} which has been studied extensively in literature
\cite{tishby08,koller07}. In fact, feature classes are a special case
of {\it meta - features} when the feature has only one meta attribute,
as gene classes or topic of the word etc. in our setting. 

More generically, starting from any set of features, one can generate new
classes of features by using projections such as principle components
analysis (PCA) or non-negative matrix factorization (NNMF),
transformations such as log or square root, and interactions (products of features). 
Further ``synthetic'' feature classes can be created by finding clusters (e.g., using k-means) in the feature 
space as show later in the experiments section. 

Firstly we describe the notation and then we present the TPC
scheme and compare it with standard RIC coding \cite{FosterGeorge}.
Then we present an algorithm for doing ``joint'' feature selection using TPC.

\section {Notation Used}
The symbols used throughout the rest of this section are defined in the following Table \ref{tab:SymbolsTPC}:
\begin{table} [htbp]
\setlength{\abovecaptionskip}{0pt} 
\setlength{\belowcaptionskip}{0pt}
\caption{Symbols used and their definitions.}
\centering
\begin{small}
\label{tab:SymbolsTPC}
\begin{center}
\begin{tabular}{cl}
\hline
Symbol & Meaning\\
\hline
\small
n & Number of observations\\
m &  Number of candidate features\\
$m^*$ & Number of beneficial features in the candidate  \\
& feature set\\
q &  Number of features currently included in the model\\
Q & Number of feature classes currently added in the\\
& model\\
K & Total number of feature classes\\
$m_k$ & Total number of candidate features in the $k^{th}$ \\
& feature class\\

\hline
\end{tabular}
\end{center}
\end{small}
\end{table}

All the above values are given by the data, except $m^*$ which is unknown, and $q$ and $Q$, which are determined by the search/optimization procedure.

\section{Coding Schemes used in TPC}\label{TPCcodingscheme}

{ \bf Coding Scheme for $\Delta S_E$ :}

$\Delta S_E$ represents the increase in likelihood of the data by adding the new feature to the model. When doing linear regression, we assume a Gaussian model and hence have:
\begin {equation}
P(e_i)= \frac{1}{\sqrt{2\pi\sigma^2}} \exp{\left(- \frac{e_i^2}{2\sigma^2}\right) }
\end {equation}
where $e_i$ is the $i^{th}$ row of the ${\bf E}$ matrix i.e. $({\bf Y} - {\bf X}\beta)$ and ${\sigma}^2$ is the variance of the Gaussian noise.

Now, we have :\\
\begin{equation}\label{likelihoodTPC}
S_E = -\log\left(\prod_{i=1}^{n}P(e_i)\right) 
\end{equation}

Note that the above equation~\ref{likelihoodTPC} is quite similar to the $S_E$ equation for MIC~\ref{negloglike}; the only difference being that here we have a single response (task).

Intuitively, $\Delta S_E$ corresponds to the increase in benefit by
adding the new feature to the model. It is always non-negative;
even a spurious feature cannot decrease the training data likelihood.

{\bf Coding Scheme for $\Delta S_M$ :}

To describe $\Delta S_M$, when a new feature is added to the model, we
use a three  part coding  scheme. Let $l_C$ be the number of
bits needed to code the index of the ``feature class" of the evaluated
feature, let $l_I$ be the number of bits used to code the index of the
evaluated feature in that particular feature class, and let $l_\theta$
be the number of bits required to code the coefficient of the
evaluated feature. Thus:

\begin {equation}\label{deltasm}
\Delta S_M = l_C + l_I + l_\theta
\end {equation}

This coding, as specified below, is the source of the power of our
approach. Intuitively, if a feature class has many good (beneficial)
features then we can share the cost of coding $l_C$ across the
features and hence save many bits in coding, as each feature costs
roughly $log(m_k)$ bits to code rather than $log(m)$ as required by
the standard RIC penalty. Soon, we will do an exact mathematical
analysis and show why this improvement occurs. But, before that we
need to explain how to code each of the three terms on right hand side
of Equation~\ref{deltasm}.

\paragraph{\textbf{Code $l_C$:}}
$l_C$ represents the number of bits required to code the index of the
feature class to which the evaluated feature belongs. When we are
doing feature selection by using TPC, two cases can arise:

{\textbf{Case 1:} The feature class of the feature being evaluated is
  not yet included in the model. In this case, we
  code $l_C$ by using $log(K)$ bits, where \textit{K} is the total
  number of feature classes in the data. From now on, we will denote
  $l_C$ under this case as $l_C^1$ .

\textbf{Case 2:} The feature class of the feature being evaluated is
already included in the model. In this case, we can save some bits by
coding $l_C$ using $log(Q)$ bits where \textit{Q} is the number of
feature classes included in the model till that point of time. (Think
of keeping an indexed list of length $Q$ of the feature classes that
have been selected.)  This is where TPC wins over other methods, as
we do not need to waste bits on coding the feature class if it is
already in the model. We will call $l_C$ under this case as $l_C^2$.

We can summarize the coding scheme for $l_C$ as follows:

\begin {equation}
l_C = \left\{ \begin{array}{ll}log(K) &  if\; the\; feature\; class\; is\; not\; in\; the \\
 & model\\ 
log(Q) & if\; the\; feature\; class\; is\; already\; in \\ 
& the\; model 
\end{array} 
\right.
\end {equation}

\paragraph{\textbf{Code $l_I$:}}
$l_I$ represents the number of bits required to code the index of the
feature within its feature class. We have a total of $m_k$ features in the
$k^{th}$ feature class. We use an RIC-style coding to code $l_I$
i.e. we use log($m_k$) bits to code the index of the feature.  (This is equivalent to
the widely used Bonferroni penalty.) Since
we also code the coefficient of the feature $l_\theta$ (unlike
standard RIC), we do not overfit even when the usual RIC assumption of
$n << m_k$ is not valid.
\begin {equation}
l_I = log(m_k)
\end {equation}

\paragraph{\textbf{Code $l_\theta$:}}

This term corresponds to the number of bits required to code the value
of the coefficient of each feature. We could use either AIC or the
more conservative BIC criterion to code the coefficients. We use 2
bits for each coefficient, which is quite similar to the AIC
criterion.
\begin {equation}
l_\theta = 2
\end {equation}
The detailed criteria for making this choice is explained in Section~\ref{AICcode}

\section{Analysis of TPC Scheme}

We now compare the TPC coding scheme with a standard coding scheme
(abbreviated as SCS below) in which we use an RIC penalty for feature
indexes and an AIC-like penalty (2 bits) for the coefficients of the
features, as this is the form of standard feature selection setting
that comes closest to TPC in theory and in performance.

The Total Cost in bits used by SCS to code the \textit{q} selected features is:

\begin{eqnarray}
\label{TPC1}
Total Cost_{SCS}  &=& \overbrace{[qlog(m)]}^{RIC \hspace {0.01 in} Penalty} + \overbrace{[2q]}^{Coefficients} \nonumber \\
            &=&qlog(K) +qlog(\frac{m} {K}) + 2q
\end{eqnarray}

The total cost used by TPC to code the same features is:

\begin{eqnarray}
\label{TPC2} 
Total Cost_{TPC}& = &\overbrace{Qlog(K)}^{l_C^1}  + \overbrace{(q - Q)log(Q)}^{l_C^2}\nonumber\\
& & + \overbrace{qlog(m_k)}^{l_I} + \overbrace{2q}^{l_\theta}
\end{eqnarray}

The savings in coding comes from the (q-Q) features that belonged to classes that were already in the model.

\paragraph {\textbf{Case 1: All Classes are of uniform size:}}

In this case, log($m_k$) in the Equation \ref{TPC2} will be equal to
log($\frac {m} {K}$), as the size of each feature class will be same
and will be equal to $\frac {m} {K}$, where \textit{m} is the total
number of candidate features and \textit{K} is the total number of
feature classes. So, subtracting Eq.~(\ref{TPC2}) from Eq.~(\ref{TPC1}) we get:

\begin{equation}\label{netcost}
\Delta Total Cost = (q-Q)log(\frac{K} {Q})
\end{equation}

Equation~\ref{netcost} shows that TPC gives substantial improvement
over SCS when either one or both of the conditions $q \gg Q$ or $K \gg
Q$ are true. In other words, TPC wins when there are more features $q$
than feature classes $Q$ included in the model (i.e., there are
multiple features per class) or, a smaller fraction,
\textit{Q}/\textit{K}, of the feature classes include selected
features.

In short, the real performance gain of TPC occurs when all or most of
the (beneficial) selected features lie in small number of feature
classes. The best case would occur when all the (beneficial) selected
features lie in one class and the worst case occurs when the
beneficial features are uniformly distributed across all the feature
classes.
In real datasets, the scenarios that we encounter lie somewhere
between the best and the worst case, so we can expect substantial
performance gain by using TPC.

\paragraph {\textbf{Case 2: Classes are of nonuniform size:}}

In this case, much of the theory remains the same as in Case 1, except
that $C \neq \frac{m}{K}$. Let $\frac{m}{K}$ = $m_{avg}$, i.e., the
average size of a feature class. Then equation \ref{netcost}
becomes:
\begin{equation}\label{case2}
\Delta TotalCost = (q-Q)log(\frac{K} {Q}) + \overbrace{q log(\frac{m_{avg}}{m_k}}^{Term 2})
\end {equation}

Now, it can easily be inferred that $m_{avg}$ $>$ $m_k$ occurs in the
case when the beneficial features are in feature classes whose size is
less than the average size of a feature class. Intuitively, $m_{avg}$
= $m_k$ occurs if the size of all the feature classes is same (which
was Case 1), so the performance of TPC will be improved in this case
compared to Case 1 if the beneficial features lie in small
classes. The improvement in performance over Case 1 will be quite
significant when the beneficial features lie in a small class i.e. C
is small or there are very big classes with no beneficial features in
them, in either case the contribution of Term 2 in Equation
\ref{case2} will increase.

\section{ Algorithms for Feature Selection using TPC}

Algorithm 1 give a standard stepwise feature selection algorithm that
uses TPC coding scheme. The algorithm makes multiple passes through
the data and at each iteration adds the best feature in the model
(i.e., the feature that has the maximum $\Delta S$). It stops when no
feature provides better $\Delta S$ than in the previous iteration.

\begin{algorithm}[htdp]
\small \caption{Forward Stepwise regression using TPC Scheme}
\begin{algorithmic}[1]
\STATE $flag = True$;       // flag for indicating when to stop
\STATE $model$ = \{\};      // initially no features in model
\STATE $prev\_max = 0$;     // keeps track of the value of $\Delta S_E$ in the previous iteration 
\WHILE {\{flag == True\}}  
\FOR {\{i = 1 to m\}}           
\STATE $Compute$ $\Delta S_E^i$;  // Increase in likelihood  by adding feature `i' to the model    
\STATE $Compute$ $\Delta S_M^i$;  // Number of extra bits required to code the $i^{th}$ feature
\STATE $\Delta S^i := \Delta S_E^i - \Delta S_M^i$;
\ENDFOR
\STATE $i_{max} := argmax_i\{\Delta S^i\}$; //The best feature in the current iteration
\STATE $current\_max := max_i\{\Delta S^i\}$; //The best penalized likelihood change in the current iteration
\IF {\{$current\_max > prev\_max$\}}
\STATE $model := model \bigcup \{i_{max}\}$; // Add the current feature to model
\STATE $prev\_max := current\_max$;
\ELSE
\STATE $flag := False$;
\ENDIF
\ENDWHILE

\end{algorithmic}
\end{algorithm}

It can be the case that it is not worth adding one feature from a
particular class, but it is still beneficial to add multiple features
from that class. In this case, it will be advantageous to use a mixed
forward-backward stepwise regression strategy in which one continues the
search past the stopping criterion given above, and then sequentially
removes the ``worst'' feature from the now overfit model. This slight
gain in search cost can find better solutions.
 
Another algorithm which can be used is streamwise feature selection,
which is greedier than the above stepwise regression methods, and
works well when there are millions of candidate features. In
streamwise feature selection, each feature is considered only once for
addition to the model, and added if it gives significant reduction in
penalized likelihood, or otherwise discarded and not examined again.

\section{Experimental Results}

In this section we demonstrate the results of the TPC scheme on
real datasets. For our experiments we use the  Stepwise TPC coding
scheme and compare against standard stepwise regression with an RIC penalty,
Lasso~\cite{Lasso}, Elastic Nets~\cite{elastic_net} and Group Lasso/ Multiple Kernel Learning~\cite{grouplasso}.

For Group Lasso/Multiple Kernel Learning, we used a set of 13
candidate kernels, consisting of 10 Gaussian Kernels (with bandwidths
$\sigma=0.5 - 20$) and 3 polynomial kernels (with degree 1-3) for each
feature class as is done by \cite{jmlr08}.  In the end the kernels
which have non zero weights are the ones that correspond to the
selected feature classes. Since GL/MKL minimizes a mixed $\ell_1/\ell_2$ norm
so, it zeros out some feature classes.
(Recall that GL/MKL gives no sparsity at the level of features within a
feature class). The Group Lasso\cite{grouplasso} and Multiple Kernel
Learning are equivalent, as has been mentioned in \cite{bach08},
therefore we used the {\it SimpleMKL} toolbox \cite{jmlr08} implementation for our
experiments. For Lasso and Elastic Nets we used
their standard LARS (Least Angle Regression) implementations
\cite{lars}. When running Lasso and Elastic Nets, we
pre-screened the datasets and kept only the best $\sim$ 1,000 features
(based on their p-values), as otherwise LARS is prohibitively slow. (The authors of
the code we used do similar screening, for similar reasons.)
For all our experiments on Elastic Nets \cite{elastic_net} we chose
the value of $\lambda_2$ (the weight on the $\ell_2$ penalty term), as
$10^{-6}$.

We demonstrate the results on real datasets pertaining to Word Sense
Disambiguation (WSD) \cite{WSD1} and gene expression data
\cite{GSEA}.  As is shown below, the results were quite
encouraging.

\subsection{Evaluation on Real Datasets (WSD and GSEA)}

In order to benchmark the real world performance of our TPC coding scheme, we chose two datasets pertaining to two diverse applications of feature selection methods, namely Natural Language Processing (NLP) and Gene Expression Analysis.
More information regarding the data and the experimental results are given below.
\paragraph{\textbf{Word Sense Disambiguation (WSD) Dataset:}}\label{palmerdata}

A WSD dataset consisting of 172 ambiguous verbs and a rich set of
contextual features \cite{WSD1} was chosen for evaluation. It consists
of hundreds of observations of noun-noun collocation,
noun-adjective-preposition-verb (syntactic relations in a sentence)
and noun-noun combinations (in a sentence or document).  The size of
the WSD data and other relevant information are summarized in Table
\ref{tab:wsdData}. We show results on 10 verbs picked randomly from the set of entire
172 verbs.

\begin{table}[htbp]
\setlength{\abovecaptionskip}{0pt} 
\setlength{\belowcaptionskip}{0pt}
\caption{Word Sense Disambiguation Dataset.}
\label{tab:wsdData}
\begin{center}
\begin{small}
%\begin{sc}
\begin{tabular}{lccc}
\hline

Dataset &  \# Observations &  \# Features &  \# Classes\\
\hline

acquire & 101 & 1081 & 43\\
care & 131 & 621 & 40\\
climb & 84 & 676 & 41\\
fire & 132 & 1217 & 43\\
add-1 & 320 & 2583 & 42\\
expand & 222 & 2144 & 42\\
allow & 344 & 2657 & 41\\
drive & 191 & 1584 & 43\\
identify & 102 & 964 & 43\\
promise & 111 & 929 & 41\\
\hline
\end{tabular}
%\end{sc}
\end{small}
\end{center}
\end{table}

A sample feature vector, given below, shows typical features and their
classes.  In each case, the part of the feature before the
underscore is the feature class. Classes included pos (part of speech
of the verb), morph (verb morphology), sub (the subject of the verb),
subjsyn (the wordnet synonym set labels of the subject), dobj (the
direct object of the verb), dobjsyn (dobj's wordnet synsets), word-1,
word-2, word+1, word+2 (the words 1 or 2 before the verb or 1 or 2
after) pos-1, pos-2, pos-3, pos-4 (the parts of speech of those
words), bigrams of the words, and tp (the topics of the document).
\begin{table*}[htbp]
    \caption{10 Fold CV accuracies of various methods on the WSD dataset (10 verbs).
    {\it Note:} (\#f) represents the average number of features selected over the 10 folds.
             {\bf These are true cross validation accuracies and no parameters have been tuned on them.} All the accuracies are ($L_1$) classification accuracies.}
    \label{tab:wsdResults}
    %\vskip 0.01in
  \small
  \centering
   \begin{tabular}{lccccc}\hline   
      Method & acquire & care\\ 
     & $\mu\pm\sigma$  ({\it \#f})& $\mu\pm\sigma$  ({\it \#f})\\
    \hline
Stepwise TPC & 95.1$\pm$0.7 (1.8) & {\bf97.7$\pm$0.5 (2.0)}\\

Stepwise RIC & 93.1$\pm$0.3 (5.1) &93.1$\pm$1.3 (12.3) \\

Elastic Nets & 90.5$\pm$0.2 (6.1) &96.1$\pm$0.2 (24.1) \\

Lasso & 90.0$\pm$0.4 (15.2) &85.4$\pm$0.9 (35.8)\\

Group Lasso/MKL & {\bf 96.0$\pm$0.1 (50.3)} & 96.0$\pm$0.3 (21.7)\\

\hline
 Method & expand & allow\\ 
    & $\mu\pm\sigma$  ({\it \#f})& $\mu\pm\sigma$  ({\it \#f})\\
    \hline
Stepwise TPC  & {\bf 99.5$\pm$0.4 (1.8)} & {\bf 95.6$\pm$1.1 (4.0)}\\

Stepwise RIC  &96.4$\pm$0.4 (4.3) &88.5$\pm$3.1 (22.4)\\

Elastic Nets &99.1$\pm$0.3 (106.9) & 93.9$\pm$0.9 (5.8) \\

Lasso & {\bf 99.5$\pm$0.3 (81.9)} & 89.1$\pm$1.0 (69.9)\\

Group Lasso/MKL & 97.7$\pm$0.7 (53) & 92.6$\pm$2.3 (2294) \\

\hline

 Method & fire & add-1\\ 
     & $\mu\pm\sigma$  ({\it \#f})& $\mu\pm\sigma$  ({\it \#f})\\
    \hline
  
Stepwise TPC & {\bf 99.2$\pm$0.6 (1.9)} &{\bf 96.6$\pm$0.4 (4.3)}\\

Stepwise RIC &95.5$\pm$1.4 (3) &91.9$\pm$2.3 (17.2)\\

Elastic Nets &95.4$\pm$1.2 (107.8) & 93.4$\pm$0.5 (1)\\

Lasso & 93.1$\pm$1.3 (106.7) & 93.8$\pm$0.2 (1) \\

Group Lasso/MKL & 97.5$\pm$0.3 (12) & 91.3$\pm$1.5 (1952)\\

\hline

Method  &identify & promise\\ 
     & $\mu\pm\sigma$  ({\it \#f})  & $\mu\pm\sigma$  ({\it \#f})\\
    \hline
  
Stepwise TPC & {\bf 99.0$\pm$0.2 (2.2)} &{\bf 96.4$\pm$0.5 (3.0)}\\

Stepwise RIC &{\bf 99.0$\pm$0.5 (1.9)}&88.2$\pm$3.1 (6.4)\\

Elastic Nets  & 89.0$\pm$1.1 (41.2)&91.9$\pm$1.7 (4.8)\\

Lasso  & 86.0$\pm$0.9 (10)& 88.3$\pm$1.1 (20.6)\\

Group Lasso/MKL  & 97.3$\pm$0.6 (1)& 90.4$\pm$1.2 (232)\\

\hline
      Method  & climb & drive\\ 
     & $\mu\pm\sigma$  ({\it \#f})  & $\mu\pm\sigma$  ({\it \#f})\\
    \hline

Stepwise TPC & {\bf 98.8$\pm$0.7 (1.9)} &{\bf 99.0$\pm$0.3 (1.4)}\\

Stepwise RIC  &88.8$\pm$3.6 (3.7) &92.1$\pm$3.1 (6.0) \\

Elastic Nets  &92.5$\pm$1.1 (91) &96.9$\pm$1.0 (1.3) \\

Lasso  & 88.8$\pm$1.1 (84.9)  & 92.1$\pm$1.4 (18.1) \\

Group Lasso/MKL  & 95.9$\pm$0.6 (11) & 97.5$\pm$0.4 (28) \\

\hline

\end{tabular}
%\vskip 0.01in 
\end{table*}
The results for the WSD Dataset are presented in the Table~\ref{tab:wsdResults}.
They show that the number of features selected vary --
sometimes the TPC select more features than other methods and vice
versa -- but the classification accuracy for TPC is higher than other methods, in $7$ out of $10$ cases.
It is equal to the accuracy of the best method on $2$ occasions and once it is slightly worse than GL/MKL.
Overall, on the entire set of $172$ verbs TPC is significantly (5 \% significance level (Paired t-Test)) better than the competing methods on $160/172$ verbs and has the same accuracy as the best method on $4$ occasions.

The accuracies averaged over all the $172$ verbs\footnote{Note: These accuracies are for the (1 vs all) 2 class prediction problem i.e. predicting the most frequent sense. On the other hand the accuracies as given in Section~\ref{TransTPC} are for the multi-class problem where we want to predict the exact sense.} are shown in Table~\ref{accuracyAll}.

\begin{table}[htbp]
\setlength{\abovecaptionskip}{0pt} 
\setlength{\belowcaptionskip}{0pt}
\caption{ $10$ Fold CV  accuracies averaged over 172 verbs}
\label{accuracyAll}
\begin{center}
\begin{small}
%\begin{sc}
\begin{tabular}{ccccc}
\hline

Stepwise TPC  & Stepwise RIC & Elastic Nets & Lasso & Group Lasso/MKL\\ 
    \hline
89.81\% & 84.19\%& 86.29\%& 85.94\% & 87.63\% \\

\hline
\end{tabular}
%\end{sc}
\end{small}
\end{center}
\end{table}

\paragraph{\textbf{Gene Set Enrichment Analysis (GSEA) Datasets:}} 

The second real datasets that we used for our experiments were gene
expression datasets from GSEA \cite{GSEA}. There are multiple
gene expression datasets and multiple criteria on which the genes can
be grouped into classes. For example, different ways of generated gene
classes include C1: Positional Gene Sets, C2: Curated Gene Sets, C3:
Motif Gene Sets, C4: Computational Gene Sets, C5: GO Gene Sets.

For our experiments, we used gene classes from the C1 and C2
collections. The gene sets in collection C1 consists of genes belonging
to the entire human chromosome, divided into each cytogenetic band that has
at least one gene. Collection C2 contained gene sets from various
sources such as online pathway databases and knowledge of domain
experts.

The datasets that we used and their specifications are as shown in Table~\ref{tab:GSEAData}.
\begin{table}[htbp]
\setlength{\abovecaptionskip}{0pt} 
\setlength{\belowcaptionskip}{0pt}
\caption{GSEA Datasets.}
\label{tab:GSEAData}
\begin{center}
\begin{small}
%\begin{sc}
\begin{tabular}{lccc}
\hline

Dataset &  \# Observations &  \# Features &  \# Classes\\
\hline

leukemia (C1) & 48 & 10056 & 182\\
gender 1 (C1) & 32 & 15056 & 212\\
diabetes (C2) & 34 & 15056 & 318\\
gender 2 (C2) & 32 & 15056 & 318\\
p53 (C2) & 50 & 10100 & 308\\
\hline
\end{tabular}
%\end{sc}
\end{small}
\end{center}
\end{table}

The results for these GSEA datasets are as shown in the Table~\ref{GSEA1} below:
\begin{table*}[htbp]
    \caption{10 Fold CV accuracies of various methods on the GSEA datasets.
	 {\it Note:} (\#f) represents the average number of features selected over the 10 folds.
            {\bf These are true cross validation accuracies and no parameters have been tuned on them.} All the accuracies are ($L_1$) classification accuracies.}
    \label{GSEA1}
    %\vskip 0.01in
  \small
  \centering
  \begin{tabular}{lccc}\hline   

      Method & leukemia & diabetes\\ 
     & $\mu\pm\sigma$  ({\it \#f})& $\mu\pm\sigma$  ({\it \#f})\\
    \hline
Stepwise TPC & {\bf 95.8$\pm$0.8 (6.3)} & {\bf80.1$\pm$1.1 (3.7)}\\

Stepwise RIC & 87.5$\pm$1.1 (4.1) &77.3$\pm$1.6 (4.4)\\

Elastic Nets & 91.1$\pm$0.6 (7.1) &78.0$\pm$0.3 (9.1) \\

Lasso & 89.9$\pm$0.5 (15.2) &77.6$\pm$0.7 (14.8)\\

Group Lasso/MKL &  93.0$\pm$0.1 (2263) & 78.7$\pm$0.4 (7139)\\
 \hline

      Method & gender -1 & gender -2 \\ 
     & $\mu\pm\sigma$  ({\it \#f})& $\mu\pm\sigma$  ({\it \#f})\\
    \hline
  
Stepwise TPC & {\bf 93.8$\pm$0.9 (4.2)} & {\bf 96.9$\pm$1.3 (4.0)}\\

Stepwise RIC &{\bf 93.8$\pm$0.9 (4.3)} &94.5$\pm$1.2 (4.5)\\

Elastic Nets &90.6$\pm$0.4 (13.1) & 92.9$\pm$0.7 (5.8)\\

Lasso &  90.4$\pm$0.3 (16.7) & 93.1$\pm$0.5 (7.9)\\

Group Lasso/MKL  & {\bf 93.8$\pm$0.7 (5084)} & 93.4$\pm$1.4 (10150) \\

\hline
      Method & p53\\ 
     & $\mu\pm\sigma$  ({\it \#f})\\
    \hline
  
Stepwise TPC & {\bf 74.0$\pm$2.1 (1.1)}\\

Stepwise RIC &66.0$\pm$1.4 (2.1)\\

Elastic Nets &70.1$\pm$0.8 (6.8)\\

Lasso & 69.8$\pm$0.4 (7.4)\\

Group Lasso/MKL & 72.5$\pm$0.3 (5792)\\
\hline

\end{tabular}
%\vskip 0.01in 
\end{table*}

For these datasets, TPC also beat the standard methods. 
Here also TPC is significantly better than the competing methods.
It is interesting to note that TPC methods sometimes selected
substantially fewer features, but still gave better
performance than other methods. This is consistent with the predictions of Equation
\ref{netcost} in that although the number of features selected, $q$
may be small, the number of classes, $K$, is quite large for the GSEA
datasets.

\chapter{Model 3: Transfer-TPC}\label{TransTPC}
In this chapter we describe our last model, Transfer-TPC which falls in the second category of transfer learning models which do ``sequential transfer'' i.e. the task on which we want to transfer knowledge may not be known in advance but it is similar to other tasks according to some ``similarity metric''. Transfer-TPC is most beneficial when we want to transfer knowledge between tasks which have unequal amounts of labeled data. Transfer-TPC not only improves the learning on tasks which have lesser amount of data but also gives siginificant benefits in predictive accuracy on tasks which have comparable amount of data.

So, first of all we describe our transfer learning formulation, Transfer-TPC which uses TPC, as described in last chapter, to do transfer between tasks.
\section{Transfer Learning Formulation}
\label{sect:pdf}

Transfer-TPC uses TPC as described in Chapter~\ref{TPCChapter} as a baseline model.
TPC provides more accurate predictions
than competing methods and can easily be extended to incorporate prior
information and share information between similar
tasks, shown below. Priors on features and feature classes, as learned by transfer
from other ``similar'' tasks, change the cost of coding a feature or a feature
class.  The number of bits that should be used to code a fact, such as
a feature being included in the model, is the log of the probability of
that fact being true. Thus, having better estimates of how likely a
feature is to be included in the model allows more efficient
coding. Similarly, knowing how likely it is that some feature in a
given class of features will be included in the model allows us to code that feature
class more precisely.  Using priors from similar tasks to better code features and
feature classes is at the core of Transfer-TPC.

\subsection{Transfer TPC}

For Transfer-TPC, we define two binary random variables
$fc_i$ and $f_j$ $\in$ \{0,1\} that denote the events of the $i^{th}$
feature class and the $j^{th}$ feature being in or not in the
model for the test task. To be more precise, $fc_i =1$ denotes the
event of $i^{th}$ feature class being in the model and $fc_i=0$
denotes the complimentary event of this feature class not being
selected by the model. Similar conditions hold for the case of
features $f_j$. We can parameterize the distributions as follows:

\begin{eqnarray}
\label{equationdist}
&& p(fc_i=1 | \theta_i) =\theta_i \\
&& p(f_j=1 | \mu_j) = \mu_j
\end{eqnarray} 

In other words, we have a Bernoulli Distribution over the feature
classes and the features. It can be represented compactly as:

\begin{eqnarray}
&& Bernoulli(fc_i| \theta_i) =\theta_i^{fc_i}(1 - \theta_i)^{1 - fc_i} \\
&& Bernoulli(f_j| \mu_j) =\mu^{f_j}(1 - \mu_j)^{1 - f_j}
\end{eqnarray}

If we have a total of $t$ training tasks then given the data for $j^{th}$ feature for all the training tasks:  $\mathcal{D}_j =\{f_{j1},...,f_{jv},...,f_{jt}\}$; we
can construct the likelihood functions from the data (under the i.i.d assumption) as:

\begin{eqnarray}
\nonumber
&& p(\mathcal{D}_{fc_i} | \theta_i)= \prod_{u=1}^{t}p(fc_{iu} | \theta_i) = \prod_{u=1}^{t}\theta^{fc_{iu}}(1 - \theta_i)^{1-fc_{iu}} \\
\nonumber
&& p(\mathcal{D}_j | \mu_j)= \prod_{v=1}^{t}p(f_{jv} | \mu_j) = \prod_{v=1}^{t}\mu_j^{f_{jv}}(1 - \mu_j)^{1-f_{jv}}
\end{eqnarray}

Note: The total data vector for all the $m$ features can be represented as:\\ $\mathcal{D}~=~\{\mathcal{D}_1,...,\mathcal{D}_m\}$;
the feature class $\mathcal{D}_{fc_i}$ data can be derived from this data by considering the simple fact that a feature class will be selected i.e $(fc_i =1)$ if atleast one feature from that feature class has been selected, i.e. 
$\mathcal{D}_{fc_i}=\{\mathcal{D}_1,..., \mathcal{D}_{m_i}\}$, where we are assuming that $i^{th}$ feature class had features $\{1,...,m_i\}$

The posteriors can be calculated by putting a prior over the parameters $\theta_i$ and $\mu_j$ and using Bayes rule as follows:

\begin{equation}
p(\theta_i| \mathcal{D}_{fc_i})= p(\mathcal{D}_{fc_i}| \theta_i) \times p(\theta_i| a, b)
\end{equation}
where $a$ and $b$ are the hyperparameters of the Beta Prior
$(\theta_i^{a-1}(1-\theta_i)^{b-1})$ which is a conjugate prior for the
Bernoulli Distribution. Similarly we can write equation involving
$\mu_j$ for the posterior over features.

Using the posterior obtained above we can evaluate the predictive distribution of $\theta_i$ and $\mu_j$ as:
\begin{equation}
p(fc_i=1| \mathcal{D}_{fc_i})= \int_0^1p(fc_i = 1| \theta_i)p(\theta_i| \mathcal{D}_{fc_i})d\theta_i 
\end{equation}
Substituting from~\ref{equationdist} in the above equation we get:

\begin{equation}
p(fc_i=1| \mathcal{D}_{fc_i})= \int_0^1\theta_i p(\theta_i| \mathcal{D}_{fc_i})d\theta_i = \mathbb{E}[\theta_i |\mathcal{D}_{fc_i}]
\end{equation}

Similarly we can write equation for the features as:
\begin{equation}
p(f_j=1| \mathcal{D}_j)= \int_0^1\mu_j p(\mu_j| \mathcal{D}_j)d\mu_j = \mathbb{E}[\mu_j |\mathcal{D}_j]
\end{equation} 

Using the standard results for the mean and the posterior of a Beta distribution we obtain:
\begin{equation}
\label{fcequation}
p(fc_i=1| \mathcal{D}_{fc_i})= \frac{k + a}{k + l + a +b}
\end{equation}
where $k$ is the number of times that the $i^{th}$ feature class is
selected and $l$ is the complement of $k$, i.e. the number of times
the $i^{th}$ feature class is not selected in the training data. We discuss
below how to choose the hyperparameters of the beta prior, $a$
and $b$.

For the case of features also, we obtain a similar equation as:

\begin{equation}
\label{fequation}
p(f_j=1| \mathcal{D}_j)= \frac{a + c}{a + z + c +d}
\end{equation}
where $a$ is the number of times that the $j^{th}$ feature is selected and $z$ is the complement of $a$ i.e. the number of times the $j^{th}$ feature is not included in the model. As earlier $c$ and $d$ are the hyperparameters of the beta prior.

\subsection{Discussion of Transfer-TPC }

As can be seen from Equations~\ref{fcequation} and~\ref{fequation},
the probability that a feature class or a feature is selected is
a ``smoothed'' average of the number of times they were selected in the
models of tasks that are similar to the task
under consideration i.e. the training tasks ($\mathcal{D}$).
We use these probabilities to formulate a coding scheme which we call
Transfer-TPC, which incorporates the prior information about the
predictive quality of the various features and feature classes
obtained from similar tasks.

In light of the above, the coding scheme can be formulated as follows:

\begin{equation}
\label{firsteq}
S_M = -\log{(p(fc_i=1| \mathcal{D}_{fc_i}))} -\log{(p(f_j=1| \mathcal{D}_j))} + 2
\end{equation}
when that feature class has not yet been selected and,
\begin{equation}
\label{thirdeq}
S_M = \log{(Q)} -\log{(p(f_j=1| \mathcal{D}_j))} + 2
\end{equation} 
when that feature class has already been selected. 
$Q$ is the total number of feature classes that have been selected up to that point of time.

In both of the above equations, the first term codes the feature
classes, the second term represents the coding for the features, and
the third term codes the coefficients. The negative signs appear
before some quantities due to the fact that those terms are negative
since they represent $\log$ of fractional numbers. They also allow the
coding scheme to be directly compared to the standard TPC coding
scheme, as we explain shortly. The above equations are used as the
coding scheme in Setting 1 in our experiments as we explain later. For
Setting 2 of our experiments, the coding scheme is slightly different as we transfer a prior only on features; hence
Equation~\ref{firsteq} changes to:

\begin{equation}
\label{secondeq}
S_M = \log{(K)} -\log{(p(f_j=1| \mathcal{D}_{j}))} + 2
\end{equation}
where $K$ is the total number of feature classes. Besides this, the two settings are the same.
\subsection{Choice of Hyperparameters}

The hyperparameters, $a$ and $b$ in Equation~\ref{fcequation} and $c$ and
$d$ in Equation~\ref{fequation} control the ``smoothing''
of our probability estimates, i.e. how strongly we want to believe the
evidence obtained from the training data (i.e. similar word senses) and how much effect we think
it should have on the model that we learn for the test task.

In all our experiments, we set $a=1$ and set $b$ so that in the
limiting case of no transfer i.e. ($k=l=0$ in
Equation~\ref{fcequation}) the coding scheme will reduce to the
standard TPC Scheme as discussed in Section 2. Thus, we
choose $b = K-1$ where $K$ is the total number of feature classes
in the test task and similarly, we choose $c=1$ and $d=m_k-1$
where $m_k$ is the total number of features in the $k^{th}$ feature
class for the test task.

As a consequence of the above choice of the hyperparameters, in most
cases we give less weight to the prior if there are few tasks in
the training set.  I.e., if there are only one or two tasks
similar to the target test task, then the prior on the test task will
be weaker than if there were many similar tasks to transfer from.

\section{Experiments}
In this section we demonstrate the experimental results of Transfer-TPC on real Word Sense Disambiguation (WSD) data in a variety of settings. The various tasks in this case were the various senses of the different words.
Firstly, we give an overview of our algorithm i.e. how we used Transfer-TPC to the WSD problem, then
we  provide description of our data and explain the similarity metric that we used for defining similarity between different word senses.

\subsection{Overview of our algorithm}

Our transfer learning approach has several steps:

\begin{itemize}
 \item Learn a separate model for distinguishing the different senses of each word. This results in logistic regression
       models for distinguishing each sense from all other senses of that word. 
       Use feature selection so that these models have relatively small sets of features.
 \item Cluster word senses based on those features from their models that are positively correlated with those particular word senses. I.e., 
       characterize each word sense by those features in its model that have positive regression coefficients. 
       (In general, features with positive coefficients are associated with the given sense, and those with negative coefficients
       are associated with other senses of that word.) Clusters should only contain highly similar senses, so many senses will not end
       up in a cluster. We use a ``foreground-background'' clustering method that puts all singleton points into a 
      ``background cluster'', which we then ignore. 
 \item For each ``target'' word sense to be predicted, use the
   ``positive'' features of other word senses in its cluster to
   estimate the probability of the features being relevant for
   disambiguating the word that includes the target verb sense. These
   probabilities (priors) are used to specify the coding length (the log
   of the probability) when searching for the MDL model for disambiguating that word.
 \item Given models for distinguishing each word sense for a word from all other senses, disambiguate each
   occurrence of that word by choosing the sense whose model gave the highest score.
\end{itemize}

We share knowledge at the level of senses of the
words rather than at the level of words, as 
there are very few words that are similar in ``all'' their senses.  
There are, however, many words that have one or more senses that are similar to
senses of other words.  Transfer occurs in the third of the steps presented above, which uses
the models learned by other ``similar'' word senses (i.e. word senses
falling in the same cluster) to generate a prior on what features and
feature classes should be selected for the test word sense.

We show below that Transfer-TPC outperforms a variety of
state-of-the-art methods that do not do transfer learning, including
SVMs with a well-tuned kernel, TPC without transfer learning, and
simple stepwise regression with a RIC (also known as ``Bonferroni'')
penalty. We also show that transfer benefits both from sharing of
semantic features (e.g., the topic of the document the word is in) and
syntactic features (e.g., the parts of speech of the surrounding
words).  Transfer-TPC is particularly useful when transferring information from frequent to rare word senses, but gives
significant benefits even for words having similar amounts of data.

\subsection{Description of Data}
We performed our experiments on VerbNet data of 172 verbs, obtained from Martha Palmer's lab~\cite{palmer00,kipper06}.
This is the same data as was used for TPC experiments in Section~\ref{palmerdata}

All the 172 verbs had 36-43 different feature classes and a total of 1000-10000 features each.
The number of senses varied from 2 (For example ``add'') to 15 (For example ``turn''). Note that there might be some senses of the words that did not show up in the data, for example there are 3 senses of the word ``account'' according to WordNet and VerbNet but only two of them show in our data, so we disambiguate among those two senses only.

\subsection{Similarity Metric}
Finding a good similarity metric, between different word senses is perhaps one of the biggest challenges that we faced. There are many human annotated ``linguistic'' similarity lexicons, like words belonging to the same Levin classes~\cite{levin}, hypernyms or synonyms according to WordNet~\cite{wordnet1,wordnetlin} or words having the same VerbNet classes~\cite{palmer00,kipper06}. In addition to this people have used InfoMap~\begin{verbatim}(http://infomap.stanford.edu)\end{verbatim}~\cite{rainangkoller} which gives a distributional similarity score for words in the corpus . One can also do k-means or heirarchical agglomerative clustering of the word senses. But the main shortcoming of all these methods is that they allot all the word sense to some cluster, but in reality if we look at the data, there are many word senses that are not similar to any other word sense, either semantically or synactically, in such a case the distributional similarity scores returned by InfoMap mostly contain noise, and there will be a risk of fitting noise and not doing a good job of transfer on the test word sense. In essence, what we need is a similarity measure that gives us very ``tight'' clusters of word senses and does not cluster all these ``junk'' word senses which are not similar to any other word sense in the corpus.

So, to overcome these shortcomings, we use ``foreground-background'' clustering algorithm as proposed by~\cite{kandylas07}. This algorithm gives highly cohesive clusters of word senses called as ``foreground'' and puts all the ``junk'' word senses in the ``background''. It may help to think about the analogy with Computer Vision where foreground represents the region of interest and background consists of everything else.

In our setting we firstly find positively correlated features for each sense of the word separately, using the Simes' method~\cite{simes,simes2}, as these are the ``true'' features for that particular word sense.
For example for one sense of the word ``fire'' which means ``to dismiss somebody from employment '' the positive features were
\begin{verbatim}
`company', `executive', `friday', `hired', `interview', `job', 
`named',`probably',`sharon', 'wally', 'years', `join', `meet',
`replace', `pay',`quoted',....,`VBD-VBN', `VB-VP',`VB-VP-NP-PRP',
 `PRP', `VBD-NP'... etc.
\end{verbatim}
whereas for another sense of the same word  ``fire'' which means ``to ignite something'' the positive features were
\begin{verbatim}
`prehistoric', `same', `temperature',`israeli', 
`palestinian',`incident',`months', `retaliation',`showing'...
`NNP-NP-S-VP-VBD', `NNP-NP-S-VP-VBD',`NNP',`NNP-NNP-VBD',`NNP-VBD' 
... etc.
\end{verbatim}
Note that we have shown both the semantic and syntactic positive features, though due to space constraints we did not show all of them.

Then we cluster the word senses where each word sense is represented by the above ``positive feature'' vector. These features contained both the semantic (For example the features in the ``topic'' feature class) and syntactic (For example the features in the ``pos'', ``dobj'' feature class etc.) features. Only $67\%$ of all the word senses fall into foreground clusters in this setting. The sample clusters that we got using this approach included senses of words like 
\begin{verbatim} 
{`back',`stay', `walk', `step'}, 
{`kill', `capture', `arrest',`strengthen', `release'}.
\end{verbatim} As is obvious these two clusters contain words with semantic distributional similarity only, then we had clusters like 
\begin{verbatim}{`love',`promise', `turn', `wear'}, 
\end{verbatim} where we had words with semantic (For example `love' and `promise') as well as syntactic (For example `love' and `wear' (They share the features `leftpath1\_NNS-NP-S-VP-VBP', `leftpos1\_NNS' and `leftsurfpath1\_NNS-VBD) or `wear' and `turn') similarity.
 We also report results in which we perform the clustering using only semantic features  and only syntactic features. The motivation behind doing this is that it would be interesting to see which kind of features are more repsonsible for the performance improvement due to transfer learning. Sample clusters for the case of only semantic similarity include 
\begin{verbatim} 
{`beat', `strike', `attack',`support'}, {`do', `die', `save'}, 
{`agree', `approve'}, {`end', `finish'}.
\end{verbatim} For the case of only syntactic clustering, the clusters included
\begin{verbatim}
{`beat', `respond', `urge'}, {`note', `learn', `shake'}, 
{`sleep', `write-1', `write-2'}.
\end{verbatim} These are just a small set of representative clusters, besides these we had about 60-70 more clusters in each case.

\subsection{Experimental Setup}
We break down the problem of WSD from the level of words down to the level of senses, i.e. if we have 10 verbs with 4 senses each, we will break them up into $10 \times 4$ learning tasks. Such a partitioning makes total sense as its very difficult to find a good similarity metric at the level of words i.e. it is very difficult to find two words which are similar in ``all'' the senses. But if we break the problem down to the level of senses then we can definitely find two or more words which are similar in one sense. For example, the words ``fire'' and ``dismiss'' are similar in one sense only which means ``to dismiss somebody from work'', but their other senses are quite different from each other. In such a case it would make sense to have only these senses of ``fire'' and ``dismiss'' in the same cluster for doing transfer, rather than putting all their senses in the same cluster. 

Later on, when we have learned models for each of these senses separately,  we can again combine these senses and disambiguate the word as a whole. The predicted sense of the word is the sense whose model gave the highest score. So, its quite possible that for some senses of a word, we can do lots of transfer as there are many senses of other words similar to them, but for other senses of the same word there may not be many similar senses hence we will have less transfer for those senses. In the end, it turns out that words all of whose senses had similar senses in the corpus give very high performance on WSD and for other words, for which we could only find similar senses for some of its senses, there is a lesser improvement in performance over the baseline case in which we do no transfer, which seems quite intuitive.

In order to ensure fairness of comparison we adopt the same methodology of outputting the sense whose model had the highest score,  as the most probable sense, for all the methods that we compete against. Such kind of prediction in multi-class problems is commonly known as ``one vs all'' approach.

We do Transfer in four slightly varied settings to tease apart the entire method and get more information about subtle aspects of our methodology . In our main Transfer-TPC setting (Setting 1) we transfer a prior on both the features and the feature classes of the test word sense and in this setting we cluster the word senses based on ``semantic and syntactic'' similarity. Setting 2 is similar to Setting 1 except that we transfer prior only on features and not on feature classes.  The coding scheme for setting 1 is given by Equations~\ref{firsteq},~\ref{thirdeq} and the coding scheme for setting 2 is given by Equations~\ref{secondeq},~\ref{thirdeq}. As can be seen, these schemes differ only in the way they code the feature classes.

We slightly modify the above settings in order to have further insight into the linguistic aspects of the transfer. So, we transfer a prior on only the ``semantic'' features and features classes i.e. features in feature classes like ``tp'' (topic of the document) and this time the clustering of word senses was also done based on only the ``semantic'' features (Setting 3). In Setting 4 we transfer a prior on only the ``syntactic'' features and features classes i.e. features in feature classes like ``pos'', ``dobj'' etc. and the clustering of word senses was done based on only ``syntactic'' features.

\subsection{Results} 

We compare Transfer-TPC against standard TPC, Stepwise RIC and SVM with well tuned radial basis (RBF) kernel. Besides this we also compare results with a baseline majority voting algorithm which outputs the most frequent ``sense'' for the word as the most probable sense. For Standard TPC we used the same coding scheme as mentioned in Section~\ref{TPCcodingscheme}. For SVM we used the standard libSVM package~\cite{libsvm} and tried various kernels including linear, polynomial and RBF. In the end we used the RBF as it gave best performance on separate held out data. We tuned the ``gamma'' parameter of the RBF kernel using cross validation.

The results for various settings are shown in Table~\ref{TTPC}.
As is obvious from Table~\ref{TTPC}, Setting 1 i.e. the setting in which we put a prior on features as well as feature classes of the test word sense and do ``semantic + syntactic'' clustering gave the best accuracy averaged over all the 172 verbs which is significant at $5\%$ significance level (Paired t- Test). Settings 3 and 4 in which we cluster based on only ``semantic'' and ``syntactic'' features respectively,also gave significant ($5\%$ significance level in Paired t- Test) improvement in accuracy over the competing methods. But these settings performed a bit worse than Setting 1, which suggests that it is a good idea to have clusters in which the word senses have ``semantic'' as well as ``syntactic'' distributional similarity.
Also, worth noting is Setting 2, in which we put the prior on only the features of the test word, gave slightly worse performance than Setting 1, 3 and 4 which suggests that it helps to generalize across  features as well as feature classes. 

In addition to this we would like to give some examples which re -iterate the point that we made earlier i.e. transfer helps the most in cases in which the test word sense has a lot less data than the train word senses. ``kill'' had roughly $5.5 -6$ times more data than all other word senses in its cluster i.e. ``arrest'', ``capture'', ``strengthen'' etc. and in this case Transfer-TPC gave $4.61 - 9.22\%$ higher accuracies than competing methods on these three words. Also, for the case of word ``do'' which had roughly $10$ times more data than the other word senses in its cluster like ``die'' and ``save'', Transfer-TPC gave $6.11 - 8.63\%$ higher accuracies than other methods. For the case of word ``write'' which had 4 times more data than ``sleep'' transferring improved accuracy by $4.09\%$. It is worth noting that all these reported improvement in accuracies are much more than the average improvement in accuracy over the entire $172$ verbs as reported in Table 1, which explains the fact that transfer makes the biggest difference when the test words have a much lesser data as compared to train word senses, but even in cases where the words have similar amount of data we got $2.5 - 3.5\%$ increase in accuracy.

We would also like to mention the case of negative-transfer \cite{Caruana97multitasklearning} i.e. transfer actually hurt performance. There were $8$ such verbs out of $172$ where we observed this phenomenon.
\begin{table*} [htbp]
\setlength{\abovecaptionskip}{1pt} 
\setlength{\belowcaptionskip}{1pt}

\caption{Table showing 10 fold CV accuracies of various methods for various Transfer Learning settings. {\it Note:}  {\bf These are true cross validation accuracies and no parameters have been tuned on them.} In Settings 3 and 4 we did only ``semantic'' and only ``syntactic'' clustering respectively, in contrast to Settings 1,2, so the accuracies of the competing methods i.e. TPC, SVM, Stepwise RIC and Majority Vote are a bit different across the settings, because there were words whose some senses fell in ``foreground'' in Settings 1,2  but in Settings 3 and 4 ``all'' their senses fell in ``background'' and vice versa, so all the settings did not have exactly same words.  }
\centering
\begin{small}
\label{TTPC}
\begin{center}
\begin{tabular}{|c|c|c|c|c|}
\hline
\multicolumn{5}{|c|}{Setting 1:}\\
\hline
Transfer-TPC & Standard TPC & SVM & Stepwise RIC & Majority Vote\\
\hline
\small
$86.29\%$&$83.50\%$&$83.77\%$&$79.04\%$&$76.59\%$\\
\hline
\multicolumn{5}{|c|}{Setting 2:}\\
\hline
Transfer-TPC & Standard TPC & SVM & Stepwise RIC & Majority Vote\\
\hline
$85.75\%$&$83.50\%$&$83.77\%$&$79.04\%$&$76.59\%$\\
\hline
\multicolumn{5}{|c|}{Setting 3:}\\
\hline
Transfer-TPC & Standard TPC & SVM & Stepwise RIC & Majority Vote\\
 \hline
$85.11\%$&$83.09\%$&$83.44\%$&$79.27\%$&$77.14\%$\\
\hline
\multicolumn{5}{|c|}{Setting 4:}\\
\hline
Transfer-TPC & Standard TPC & SVM & Stepwise RIC & Majority Vote\\
 \hline
$85.37\%$&$83.34\%$&$83.57\%$&$79.63\%$&$77.24\%$\\
\hline
\end{tabular}
\end{center}
\end{small}
\end{table*}

\chapter{Discussion: A Unified View}

So far, we have seen all the three models in isolation. We now look
for a unified representation of the three models and explore the
connections between them.  This provides deeper insights into the
working of the models, and on how to select the best model for a given
problem.

We have presented the three methods using an information theoretic
approach, but they can be interpreted as Bayesian models by noting
that the cost of coding an event (such as a feature being in a model)
of probability $p$ is $-log(p)$. Thus, the RIC penalty of log(m)
(log of the number of candidate features) is just $ -log(p)$ where $p=
1/m$ assumes that one of the $m$ features will enter the
model. Transfer-TPC estimates the probabilty of a feature entering the
model as being the fraction of times it was used in models on similar
tasks.  MIC and TPC, roughly speaking, model the probability of a
feature being added to a model as being the fraction of features in
the feature class that have already been added to the model. As such,
they have the flavor of an empirical Bayes model, that ends up using
as a prior for the class the fraction of features added to the class.

\section{Connection between TPC and Transfer-TPC}

TPC is the basic building block on which Transfer-TPC has been built
and in the case of no transfer these two are equivalent.

The basic TPC scheme can be represented in a Bayesian way as follows:

\begin{equation}\label{unify}
 P(f_{ij}=1) = P(f_{ij}=1 |fc_j=1)*P(fc_j=1)
\end{equation}

where $P(fc_j=1)$ is the probability of the $j^{th}$ feature class
being included in the model and $P(f_{ij}=1 |fc_j=1)$ is the
probability of the $i^{th}$ feature from the $j^{th}$ feature class
being included in the model given that $j^{th}$ feature class is
already in the model.

In the case of standard TPC, $P(fc_j=1) = \frac{1}{K}$ where $K$ is
the total number of feature classes in the data and $P(f_{ij}=1
|fc_j=1)= \frac{1}{m_k}$ where $m_k$ is the total number of features
in the $k^{th}$ feature class. Replacing the probabilities by $-log()$
probabilities Equation~\ref{unify} reduces to the TPC scheme as
explained in Chapter 6.

It can easily be seen that in case of Transfer-TPC, the above equation
holds, but the values of the probabilities depend on whether those
features and feature classes have been selected in the models of other
``similar tasks''. In that case~$P(fc_j=1)$~$=$~$\frac{k+a}{k+l+a+b}$ 
and $P(f_{ij}=1 |fc_j=1)= \frac{a + c }{a + z + c + d}$ where
the symbols have the same meaning as we explained in Chapter 7.

\section{Connection between MIC and TPC}

As pointed out earlier, MIC and TPC both do ``simultaneous transfer''
and can be used for ``joint feature selection'' for a set of related
tasks which share the same set of features.  Both put coefficients
into classes, the key difference is that in MIC the coefficient class
is the set of coefficients of a single feature in all tasks, while in
TPC, each feature class has multiple features, and is specified
explicitly.

In both cases, we first code whether any feature from a class is
added, and then which features from within the class are to be
added. This has the consequence that once one feature from a class is
added, other features become much easier to add. The coding also
assures that subsequent features are increasingly easy to add. This is
similar in spirit to widely used methods of controlling false
discovery rate in the absence of feature classes
\cite{Benjamini}.

\section{Connection between MIC and Transfer-TPC}

MIC and Transfer-TPC are the most different of the pairs of methods,
as MIC does ``simultaneous transfer'' and expects all tasks to share
same set of features whereas Transfer-TPC is more flexible and can
even work in the case when the tasks have unequal amounts of data and
the task to which we want to transfer knowledge is unknown.  In our
implementation, we assume that all tasks in MIC are potentially
related, but for Transfer-TPC, we explicitly look for tasks that are
``similar'' to the target task being learnt.

Transfer-TPC does not require that different tasks have different sets
of feature values. (Unlike MIC, which does require that the tasks
share the same feature values.)  In the case in which all the
different tasks have same set of features and all tasks are assumed to
be ``similar'' to each other, there is a direct mapping between MIC
and Transfer-TPC setting, as in that case we can rewrite the $n \times
h$ matrix in the MIC problem as $h$, $n \times 1$ matrices with all
the $h$ different tasks being in the ``same cluster'' for doing
Transfer-TPC. In short, we can say that under these conditions, MIC
and Transfer-TPC settings become same, and MIC comes out as a special
case of Transfer-TPC in which we are transferring from all the $h-1$
remaining tasks.

\chapter{Conclusion}

In this thesis we presented three related ways of using Transfer
Learning to improve feature selection.  The three approaches shared
different kinds of information between tasks or feature classes, and
were based on the information theoretic Minimum Description Length
(MDL) principle. Two of the models, MIC and TPC,  do
``joint feature selection'' for a set of related prediction tasks
which share the same set of features while the third model,
Transfer-TPC, does ``sequential transfer'' between tasks which do not share observations..
Transfer-TPC is particularly useful when transferring knowledge between tasks which have unequal
amounts of labeled data.  All the three models gave accuracies 
on a set of Genomic and Word Sense Disambiguation datasets that are uniformly
as good as or better than state-of-the-art methods, often using models that are more sparse.
We also saw that under certain conditions and assumptions all the
three models are ``inter -reducible''.  Thus, depending on the
characteristics of the prediction problem at hand we can chose one of
the methods to improve the task of feature selection by transferring
knowledge.

\bibliography{dhillon_thesis}
\bibliographystyle{dhillon_thesis}

\end{document}